%% file: main.tex
\RequirePackage{fix-cm}
\documentclass[journal]{IEEEtran} 
\usepackage{mathtools}
\usepackage{capt-of}
\usepackage{multicol}
\usepackage{amsmath,amssymb,amsfonts}
\usepackage{url}

\usepackage{needspace}
\usepackage{alltt}
\usepackage{soul}
\usepackage{siunitx}
\usepackage{pbox}
\usepackage{cite}
\usepackage{algorithmic}
\usepackage{algorithm}
\usepackage{graphicx}
\usepackage{tikz}
\usetikzlibrary{arrows.meta,positioning,fit,backgrounds}
\usepackage{textcomp}
\usepackage{xcolor}
\usepackage{booktabs}
\usepackage{multirow} 
\usepackage{array} 
\usepackage{colortbl} 
\usepackage{hyperref}
\hypersetup{colorlinks=false,pdfborder={0 0 0.5},pdfborderstyle={/S/S /W 0.5},citebordercolor={0 .55 .15},linkbordercolor={.15 .35 .75},urlbordercolor={.65 .2 .65}}
\usepackage{cleveref}   
\usepackage{colortbl}

\newtheorem{remark}{Remark} 
\newtheorem{proposition}{Prop.} 

\makeatletter
\long\def\HL#1{\if@noskipsec\leavevmode\fi\begingroup\color{blue}#1\endgroup}
\makeatother
\hypersetup{pdftitle={DD-GBP: Anytime Multi-Arm Planning and Continuous Execution via Distributed Dynamic Gaussian Belief Propagation},pdfauthor={Zhao Jin, Fengyi Dai, Yuetong Yin, Yunan Wang, Wenjie Chen, Chuxiong Hu}}
\newcommand\bsh{\boldsymbol{h}}
\newcommand\bsx{\boldsymbol{x}}
\newcommand\bsz{\boldsymbol{z}}
\newcommand\bsq{\boldsymbol{q}}

\newcommand\bseta{\boldsymbol{\eta}}
\newcommand\mbX{\mathbf{X}}
\newcommand\mbZ{\mathbf{Z}}
\newcommand\mbbR{\mathbb{R}}

\makeatletter
\let\ddgbp@fs@ruled\fs@ruled
\def\fs@ruled{\ddgbp@fs@ruled
  \def\@fs@pre{\begingroup\color{black}\hrule height.8pt depth0pt \kern2pt}%
  \def\@fs@post{\kern2pt\hrule\relax\endgroup}}
\makeatother

\makeatletter
\def\@IEEEBIOskipN{1.25\baselineskip}
\makeatother

\makeatletter
\AtBeginDocument{%
  \long\def\HL#1{\if@noskipsec\leavevmode\fi\begingroup\color{black}#1\endgroup}%
}
\makeatother
\begin{document}
\title{\HL{DD-GBP: Anytime Multi-Arm Planning and Continuous Execution\\via Distributed Dynamic Gaussian Belief Propagation}}
 
\author{
    \vskip 1em
    Zhao Jin, Fengyi Dai, Yuetong Yin, Yunan Wang,\\
    Wenjie Chen, Chuxiong Hu
    \thanks{The authors are with the State Key Laboratory of Tribology in Advanced Equipment, Department of Mechanical Engineering, Tsinghua University, Beijing, 100084, China, Beijing Key Laboratory of Transformative High-end Manufacturing Equipment and Technology, Department of Mechanical Engineering, Tsinghua University, Beijing 100084, China, and also with State Key Laboratory of High-end Heavy-load Robots, Midea Group, Foshan 528300, China.

Corresponding author: Chuxiong Hu (email: \href{mailto:cxhu@tsinghua.edu.cn}{cxhu@tsinghua.edu.cn}).}
}

\maketitle

\begin{abstract}
\HL{Safely executing long-horizon multi-arm tasks online in complex dynamic environments remains challenging. We propose a new distributed dynamic Gaussian belief propagation (DD-GBP) method that directly plans joint-space trajectories through parallel local inference. A unified Gaussian-process (GP) parameterization connects adaptive planning horizons, interpolation-time factors, and continuous execution. Each robot maintains a local tree-structured factor graph and exchanges messages asynchronously over active inter-arm links. Dynamic subnet reconstruction and event-triggered updates reduce communication and inference costs while responding to moving obstacles. We also provide theoretical analyses of local inference accuracy, anytime early stopping, triggered-replanning consistency, and GP-consistent execution.}

\HL{In 50 long-horizon, multi-target four-arm scenes with moving obstacles, DD-GBP achieves $92.5\%$ collision-free all-target completion and the highest success rate among six representative baselines. Scenario ablations demonstrate long-horizon safety and robustness to communication impairment and extreme conditions; success remains $86.0\%$ under $90\%$ packet loss. Execution results further show substantially suppressed extreme velocity and acceleration peaks relative to Syn-GBP. On one workstation, obstacle-free tests with 30 coexisting arms retain $99.87\%$ success and a $15.80$~ms mean warm solve, exceeding 60~Hz and outperforming the compared methods in scalability. Dual-arm transport experiments under dynamic interference demonstrate online response and practical deployability. To the best of our knowledge, these results establish a new state of the art for online long-horizon multi-arm planning in dynamic scenes.}
\end{abstract}


\markboth{DD-GBP: ONLINE MULTI-ARM PLANNING}%
{}

\definecolor{limegreen}{rgb}{0.2, 0.8, 0.2}
\definecolor{forestgreen}{rgb}{0.13, 0.55, 0.13}
\definecolor{greenhtml}{rgb}{0.0, 0.5, 0.0}

\section{Introduction}
\label{sec:introduction}
\IEEEPARstart{M}{ulti-manipulator} collaboration \HL{improves flexibility and productivity} in cooperative handling~\cite{park2024constrained}, logistics~\cite{WANG2025102849, zhou2019real}, and manufacturing~\cite{deWitte2022learning, dong2024bimanual, zhao2025adaptive}\HL{. In confined workspaces without physical separation, each arm's motion changes the space available to its neighbors, so collision avoidance requires coordination throughout the task. The planner must maintain this coordination while responding online to moving obstacles and disturbances, making fast solution updates and safety over long-horizon execution essential.}

\HL{Centralized planners jointly optimize all} robots\HL{' trajectories and constraints}~\cite{wagner2011m, chen2025cimap, solovey2016finding}\HL{, which allows interactions to be considered within one planning problem. Their computational cost, however, grows with the coupled state space and} can \HL{limit online multi-arm planning~}\HL{\cite{JIN2026103923}}\HL{. Distributed planners reduce this burden by assigning a local planning problem to each robot, while using} inter-robot \HL{communication} to \HL{coordinate the resulting trajectories}~\cite{zhang2023online, WEI2025102850}. \HL{Insufficient coordination can still cause collisions in dense workspaces.}

\HL{To coordinate coupled trajectories, optimization-based} distributed methods \HL{decompose planning into per-robot subproblems. The nonlinear variant (NADMM) of the alternating direction method of multipliers (ADMM)~\cite{boyd2011distributed,ferrantiDistributedNonlinearTrajectory2023} coordinates these} subproblems through consensus constraints \HL{and repeated information exchange}. For high-degree-of-freedom (DoF) manipulators, \HL{collision constraints couple many joint variables and make the local} subproblems highly nonconvex\HL{, complicating convergence and exposing poor local minima. For online operation, model predictive control (MPC) repeatedly optimizes a finite trajectory horizon, executes its first part, and updates the plan using new observations.} Distributed MPC (DMPC)~\HL{\cite{schwartingParallelAutonomyAutomated2017,ferrantiDistributedNonlinearTrajectory2023}} \HL{applies this procedure to coordinated per-robot problems, balancing horizon length against the cost of repeated solves. Short horizons reduce each update's computation} but \HL{can restrict coordination over longer motions, motivating solvers that exploit trajectory structure} to \HL{make frequent distributed updates affordable.} \HL{For manipulators, Gafur et al.~\cite{gafur2022dynamic} combine distributed nonlinear MPC (NMPC) with dynamic collision and deadlock avoidance, providing a directly relevant constrained-optimization baseline.}

\HL{A different way to reduce} online \HL{computation is} to \HL{learn planning policies from data. Learning-based planners have been demonstrated} in \HL{grid-based} multi-agent pathfinding~\cite{jiangDeployingTenThousand} \HL{and dual-arm operation~}\HL{\cite{aljalbout2022clascoordinatingmultirobot, lee2025multiagent}}\HL{, with a graph neural network (GNN)-based planner coordinating up to eight arms in static scenes}~\cite{laiRoboBalletPlanningMultirobot2025}\HL{. Their performance in new task distributions depends on how well the training data and learned policies generalize} to \HL{the new geometry. Model predictive path integral (MPPI) planning offers another route to frequent online updates. MR-STORM~}\HL{\cite{evron2026mrstorm}} \HL{uses graphics processing unit (GPU)-accelerated sampling to extend predictive control to decoupled multi-arm planning, evaluating candidate motions against other arms' predicted trajectories.}

\HL{Efficient computational structures and mature optimizers also accelerate planning. GPMP2~\cite{mukadamContinuoustimeGaussianProcess2018} uses continuous GP trajectory parameterization to construct sparse Bayesian factor graphs for single-arm maximum a posteriori (MAP) optimization with GTSAM~\cite{dellaertFactorGraphsExploiting2021}. Centralized extensions explore robot formations~\cite{guo2021formation} and 2D multi-robot scenarios~\cite{wang2022multimanipulator}; high-dimensional coupling and local minima remain challenges for centralized optimization. We likewise use GP trajectory parameterization, but go beyond sparse representation by further exploiting its advantages for asynchronous multi-arm coordination, collision queries, adaptation, and consistent execution. This representation naturally matches our local Gaussian message passing.}

\HL{For continuously updated coordination,} Gaussian Belief Propagation (GBP) \HL{provides} a distributed inference \HL{formulation over continuous states through} Gaussian graphical models~\HL{\cite{bishop2006pattern,suConvergenceConditionsGaussian2015,duConvergenceAnalysisDistributed2018,ortizVisualIntroductionGaussian2021}}\HL{. Its local message structure allows neighboring} variables \HL{to exchange information without global consensus variables or task-specific training, and} has \HL{supported localization} \HL{of planar automated guided vehicle (AGV) teams}~\HL{\cite{davisonFutureMapping2Gaussian2022,muraiRobotWebDistributed2024}} \HL{and collaborative planning}~\cite{patwardhanDistributingCollaborativeMultiRobot2023}. \HL{For manipulators, GP-based replanning by Bari et al.~\cite{bari2023fastrelinearization} remains single-arm, whereas Prados et al.~\cite{prados2025distributed} coordinate learned short-horizon policies rather than joint-space trajectories. Direct multi-arm planning further requires coupled joint motions, adaptive horizons and interaction neighborhoods, and a consistent inference--execution representation. Our Syn-GBP study~\cite{jin2026distributedSafetyAware} explored short-horizon dual-arm assembly; extending it to scalable multi-arm planning calls for a new architecture and accompanying theoretical analysis.}

\HL{Building on but differing from that preliminary study, we propose DD-GBP as a more complete algorithmic and theoretical framework. Its tree-directed local solve, dynamic interaction subnets, adaptive horizons, and common GP execution model target scalable multi-arm coordination and safe long-horizon execution; the experiments verify substantial performance gains. The new contributions are as follows.}

\begin{enumerate}

\item \textbf{\HL{Adaptive factor-graph planning with replanning consistency.}} We \HL{formulate multi-arm trajectory planning as MAP inference over} a \HL{changing factor graph, with geometry-aware collision factors and adaptive planning horizons. We prove} that \HL{pruning inactive inter-arm factors preserves the complete gated objective (Prop.~}\HL{\ref{prop:gated_equivalence}}\HL{) and establish a local bound on the change from the time-aligned previous trajectory after a replanning trigger (Prop.~}\HL{\ref{prop:replanning_consistency}}\HL{). This links online network adaptation} to a \HL{well-defined objective and a quantitative measure} of \HL{replanning consistency}.

\item \textbf{\HL{Distributed multi-arm network scheduling and anytime planning.}} We \HL{design a network update procedure that combines event-triggered subnet selection, local backward--forward inference, and asynchronous inter-arm message exchange. Each robot solves its fixed linearized local tree and communicates only on active links. We derive residual-based early-stopping bounds (Prop.~}\HL{\ref{prop:anytime_bound}}\HL{), giving deadline-limited updates} a \HL{quantitative inference-error criterion while keeping coordination localized as the team grows}.

\item \textbf{\HL{Consistent continuous-trajectory optimization and execution.}} \HL{We optimize joint-space trajectories through GP support states, whose adaptive process scale reflects the acceleration-driven motion amplitude, and use the same parameterization for interpolation-time collision factors, timestamp-aligned inter-arm queries, and execution references. We establish the connection between planning and execution by showing that, for the adopted prior, the conditional GP mean is the cubic Hermite trajectory sent to the tracking controller (Prop.}~\HL{\ref{prop:gp_execution}}\HL{). This makes continuous trajectory evaluation part of the} planning \HL{model and preserves that model when robots exchange and execute their plans}.
\end{enumerate}

\HL{To our knowledge, we provide the first matched evaluation spanning six representative multi-arm planning baselines: NADMM~\cite{ferrantiDistributedNonlinearTrajectory2023}, DMPC~\cite{schwartingParallelAutonomyAutomated2017,ferrantiDistributedNonlinearTrajectory2023}, GPMP2~\cite{mukadamContinuoustimeGaussianProcess2018}, Syn-GBP~\cite{jin2026distributedSafetyAware}, MR-STORM~\cite{evron2026mrstorm}, and Gafur's NMPC~\cite{gafur2022dynamic}. DD-GBP establishes a new state of the art among these evaluated baselines, with the highest success rate in the matched four-arm benchmark and the strongest observed scalability. We further examine mechanism and scenario ablations and physical deployment on dual Franka Emika Panda arms.}

\section{Preliminaries on Gaussian Belief Propagation} \label{sec:Gaussian_Belief_Propagation}

\subsection{\HL{Inference Problem and Gaussian Belief Propagation}}
\label{sec:Factor_Graph_Representation}
\label{sec:Relation_to_MPC}

\HL{Let} \HL{$\mbX=\{\bsx^i\}_{i=1}^{n}$} \HL{denote the unknown variables and} \HL{$\mbZ=\{\bsz_j\}_{j=1}^{m}$} \HL{the observations. The inference problem} is \HL{to determine the distribution of} \HL{$\mbX$} \HL{given} \HL{$\mbZ$}\HL{. A factor graph expresses the conditional-independence structure of the model by writing the} posterior \HL{as a product} of \HL{local factors}~\cite{bishop2006pattern}\HL{,}
\HL{\begin{equation}
p(\mbX\mid\mbZ)=\frac{1}{L}\prod_{j=1}^{m}f_j(\mbX_j,\bsz_j),
\label{eq:posterior_factorization}
\end{equation}}

\HL{where} \HL{$L$} \HL{is the normalization constant and} \HL{$\mbX_j\subseteq\mbX$} \HL{contains the variables involved in factor} \HL{$f_j$}\HL{. For Gaussian measurement errors or quadratic penalty models, the factors take the residual form}
\HL{\begin{equation}
f_j(\mbX_j,\bsz_j)=K_j\exp\!\left(-\tfrac12\|\bsz_j-\bsh_j(\mbX_j)\|_{\boldsymbol{\Lambda}_j}^{2}\right),
\label{eq:gaussian_residual_factor}
\end{equation}}
\HL{where} \HL{$\|\mathbf r\|_{\boldsymbol{\Lambda}_j}^{2}=\mathbf r^\top\boldsymbol{\Lambda}_j\mathbf r$}\HL{,} \HL{$K_j>0$}\HL{,} \HL{$\bsz_j,\bsh_j(\mbX_j)\in\mbbR^{N_j}$}\HL{, and} \HL{$\boldsymbol{\Lambda}_j\in\mbbR^{N_j\times N_j}$} \HL{is a fixed precision matrix. Belief propagation (BP) uses this} factorization \HL{to infer marginal distributions through local message passing. For continuous variables, its marginalization integrals generally have no closed form. When} \HL{$\bsh_j$} \HL{is affine, the residual factors are Gaussian in the state variables, and Gaussian belief propagation (GBP) evaluates the message products and marginalizations analytically~}\HL{\cite{ortizVisualIntroductionGaussian2021}}\HL{. Nonlinear functions are locally linearized to obtain the same Gaussian form, as described in Sec.~}\HL{\ref{sec:Factor_Linearization}}\HL{.}

\HL{Taking the negative logarithm gives} \HL{$-\log p=\log L-\sum_j\log f_j$}\HL{. Collecting terms independent} of \HL{$\mbX$} \HL{into} \HL{$C=\log L-\sum_j\log K_j$} \HL{yields}
\HL{\begin{equation}
\begin{aligned}
-\log p(\mbX\mid\mbZ)&=C+E(\mbX),\\
E(\mbX)&=\frac12\sum_{j=1}^{m}\|\bsz_j-\bsh_j(\mbX_j)\|_{\boldsymbol{\Lambda}_j}^{2}.
\end{aligned}
\label{eq:Negative_Log_Likelihood}
\end{equation}}
\HL{Thus, maximizing the posterior is equivalent to minimizing its negative logarithm, giving the maximum} a posteriori (MAP) \HL{solution}
\HL{\begin{equation}
\mbX^\star=\arg\max_{\mbX}p(\mbX\mid\mbZ)=\arg\min_{\mbX}E(\mbX).
\label{eq:map_equivalence}
\end{equation}}

\HL{For each Gaussian posterior with a positive-definite precision matrix, the mean is its unique MAP solution. After factor linearization, the corresponding weighted residual objective is quadratic, which connects the inference problem to quadratic penalty optimization. GBP nevertheless solves this model through Gaussian products and marginalizations, so an individual message update is not equivalent to a gradient-descent step. On a tree, these operations recover the exact marginal means and covariances. These are general properties of Gaussian inference. For the proposed DD-GBP method, when the coupled message iteration satisfies the convergence conditions discussed in Sec.~}\HL{\ref{sec:Convergence_and_Optimality_of_GBP}}\HL{, its fixed-point means recover the MAP solution of the current linearized Gaussian model. These properties support using the inferred means as trajectory estimates for each Gaussian planning model and publishing the latest means as anytime trajectories.}

\subsection{\HL{Message Passing and Updates on Factor Graphs}}
\label{sec:Messages_Passing_and_Updating}

\begin{figure}[t]
\centering
\begin{tikzpicture}[
  varnode/.style={circle,draw,minimum size=8mm,inner sep=0pt},
  factnode/.style={rectangle,draw,fill=black!8,minimum size=6mm,inner sep=0pt},
  every node/.style={font=\small}]
\node[varnode] (x1) at (0,0) {$\bsx^1$};
\node[factnode] (f1) at (1.35,0) {$f_1$};
\node[varnode] (x2) at (2.7,0) {$\bsx^2$};
\node[factnode] (f2) at (4.05,0) {$f_2$};
\node[varnode] (x3) at (5.4,0) {$\bsx^3$};
\node[factnode] (f0) at (0,1.1) {$f_0$};
\draw (f0)--(x1)--(f1)--(x2)--(f2)--(x3);
\end{tikzpicture}
\caption{\HL{Factor tree for} \HL{$f(\bsx^1,\bsx^2,\bsx^3)\propto f_0(\bsx^1)\,f_1(\bsx^1,\bsx^2)\,f_2(\bsx^2,\bsx^3)$}\HL{. Circles denote variable nodes and squares denote factor nodes. Messages pass in both directions along each edge.}}
\label{fig:generic_factor_tree}
\end{figure}
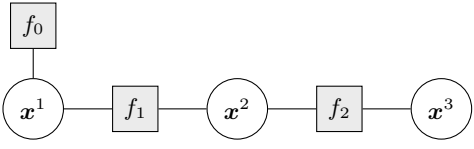

\HL{A factor graph makes the factorization in~}\HL{\eqref{eq:posterior_factorization}} \HL{explicit as} a \HL{bipartite graph. A circular variable node} \HL{$v^i$} \HL{represents} \HL{$\bsx^i\in\mbbR^{N_{\bsx^i}}$}\HL{, and a square} factor \HL{node} \HL{$w^j$} \HL{represents} \HL{$f_j$}\HL{. An edge connects} \HL{$v^i$} \HL{to} \HL{$w^j$} \HL{precisely when} \HL{$\bsx^i\in\mbX_j$}\HL{. We also use} \HL{$\mbX_j$} \HL{for its stacked vector, with dimension} \HL{$N_{\mbX_j}=\sum_{\bsx^i\in\mbX_j}N_{\bsx^i}$}\HL{. Fig.~}\HL{\ref{fig:generic_factor_tree}} \HL{illustrates a simple tree with unary and pairwise factors.}

\HL{Inference passes} messages \HL{along edges and updates} variable beliefs\HL{, with local relinearization for nonlinear} factors. \HL{We omit normalization constants and use Gaussian information form} \HL{$\mathcal{N}^{-1}(\bsx\mid\boldsymbol{\eta},\boldsymbol{\Lambda})$}\HL{, where} \HL{$\boldsymbol{\Lambda}=\boldsymbol{\Sigma}^{-1}$} \HL{and} \HL{$\boldsymbol{\eta}=\boldsymbol{\Lambda}\boldsymbol{\mu}$} \HL{for mean} \HL{$\boldsymbol{\mu}$} \HL{and covariance} \HL{$\boldsymbol{\Sigma}$}\HL{.} The \HL{superscript} \HL{$-1$} \HL{denotes information form, not a reciprocal} density\HL{. Each node maintains} \HL{$b(\bsx^i)=\mathcal{N}^{-1}(\bsx^i\mid\bseta_i,\boldsymbol{\Lambda}_i)$}\HL{. Below, the given} \HL{$\bsz_j$} is \HL{omitted from the arguments} of \HL{$f_j$}\HL{.}
\subsubsection{Messages from Variable Nodes to Factor Nodes} \label{sec:Messages_from_Variable_Nodes_to_Factor_Nodes}

The message \HL{$m_{v^i\to w^j}(\bsx^i)$} \HL{is the product of incoming messages except that} from \HL{$w^j$}\HL{, with information form}
\HL{\begin{equation}
 \mathcal{N}^{-1}(\bsx^i | \sum_{w^k \in N(v^i) \setminus w^j} \bseta_{w^k \to v^i}, \sum_{w^k \in N(v^i) \setminus w^j} \boldsymbol{\Lambda}_{w^k \to v^i}),
\label{eq:Messages_from_Variable_Nodes_to_Factor_Nodes}
\end{equation}}
\HL{where} \HL{$N(v^i)$} \HL{is the neighboring} factor set\HL{, and} \HL{$\bseta_{w^k\to v^i}$} \HL{and} \HL{$\boldsymbol{\Lambda}_{w^k\to v^i}$} \HL{are the information parameters} of \HL{$m_{w^k\to v^i}$}.

\subsubsection{Messages from Factor Nodes to Variable Nodes} \label{sec:Messages_from_Factor_Nodes_to_Variable_Nodes}

The message \HL{$m_{w^j\to v^i}(\bsx^i)$} from a factor node $w^j$ to a variable node $v^i$ is computed \HL{by marginalization, giving}

\begin{equation}
\int \cdots \int f_j(\mbX_{j})\prod_{v^k \in N(w^j) \setminus v^i} m_{v^k \to w^j}(\bsx_k) \,\mathrm{d}(\mbX_{j}\setminus\bsx^i),
\label{eq:Messages_from_Factor_Nodes_to_Variable_Nodes}
\end{equation}
where $N(w^j)$ is the \HL{neighboring variable} set\HL{, and the integral marginalizes all variables except} \HL{$\bsx^i$}\HL{. For affine} \HL{$\bsh_j$}\HL{, including the local affine approximation} of \HL{a nonlinear function in Sec.~}\HL{\ref{sec:Factor_Linearization}}\HL{, the factor} \HL{$\mathcal{N}^{-1}(\mbX_j\mid\bseta_j,\boldsymbol{\Lambda}_j)$} \HL{gives a Gaussian message with information parameters}
\HL{\begin{equation}
\begin{aligned}
\bseta_{w^j \to v^i}    & = \bseta_j^i {-} \boldsymbol{\Lambda}_j^{i, \neg i} \boldsymbol{\Lambda}_{\text{s}}^{-1} \left(\bseta_j^{\neg i} {+} \sum_{v^k \in N(w^j)\setminus v^i} \bseta_{v^k \to w^j}\right) , \\
\boldsymbol{\Lambda}_{w^j \to v^i} & = \boldsymbol{\Lambda}_j^{i,i} {-} \boldsymbol{\Lambda}_j^{i, \neg i} \boldsymbol{\Lambda}_{\text{s}}^{-1} (\boldsymbol{\Lambda}_j^{i, \neg i})^\top ,
\end{aligned}
\label{eq:Messages_from_Factor_Nodes_to_Variable_Nodes_Linear}
\end{equation}}
\HL{where} \HL{$\bseta_j {=} [\bseta_j^i; \bseta_j^{\neg i}] \in \mbbR^{N_{\mbX_j}}$} \HL{and}
\HL{$\boldsymbol{\Lambda}_j {=} \begin{bmatrix} \boldsymbol{\Lambda}_j^{i,i} & \boldsymbol{\Lambda}_j^{i,\neg i} \\ (\boldsymbol{\Lambda}_j^{i,\neg i})^\top & \boldsymbol{\Lambda}_j^{\neg i,\neg i} \end{bmatrix} $}
\HL{are the information vector and precision matrix,} \HL{$\boldsymbol{\Lambda}_{\text{s}} {=} \boldsymbol{\Lambda}_j^{\neg i, \neg i} + \sum_{v^k \in N(w^j)\setminus v^i} \boldsymbol{\Lambda}_{v^k \to w^j}$}\HL{, and} \HL{$i$}\HL{,} \HL{$\neg i$} \HL{denote the parts corresponding to} \HL{$\bsx^i$} \HL{and the remaining variables in} \HL{$\mbX_j$}\HL{, respectively. Incoming information matrices are embedded in the corresponding blocks of the stacked} variable \HL{space before addition.}

\subsubsection{Variable Belief Update} \label{sec:Variable_Belief_Update}

The variable belief \HL{is updated by taking the product} of \HL{all incoming factor} messages\HL{, giving} \HL{$\bseta_i=\sum_{w^j\in N(v^i)}\bseta_{w^j\to v^i}$} \HL{and} \HL{$\boldsymbol{\Lambda}_i=\sum_{w^j\in N(v^i)}\boldsymbol{\Lambda}_{w^j\to v^i}$}\HL{. This belief is the Gaussian marginal estimate for} \HL{$\bsx^i$}\HL{, with mean} \HL{$\boldsymbol\mu_i=\boldsymbol\Lambda_i^{-1}\bseta_i$} \HL{and covariance} \HL{$\boldsymbol\Sigma_i=\boldsymbol\Lambda_i^{-1}$} \HL{available directly for state estimation and subsequent inference.}

\subsubsection{Factor Linearization} \label{sec:Factor_Linearization}
\HL{At the current belief means} \HL{$\mbX_j^0$}\HL{, linearize the nonlinear measurement function as} \HL{$\bsh_j(\mbX_j)\approx\bsh_j(\mbX_j^0)+J_j(\mbX_j-\mbX_j^0)$}\HL{, where} \HL{$J_j$} \HL{is its Jacobian. The resulting Gaussian factor has information parameters}
\HL{\begin{equation}
\begin{aligned}
 \bseta_j^{\prime}     & = J_j^\top \boldsymbol{\Lambda}_j \big(J_j \mbX_j^0 + \bsz_j - \bsh_j(\mbX_j^0)\big) \in \mbbR^{N_{\mbX_j}}, \\
 \boldsymbol{\Lambda}_j^{\prime} & = J_j^\top \boldsymbol{\Lambda}_j J_j \in \mbbR^{N_{\mbX_j}\times N_{\mbX_j}}. 
\end{aligned}
\label{eq:Factor_Linearization}
\end{equation}}

\HL{Substituting this Gaussian factor into~}\HL{\eqref{eq:Messages_from_Factor_Nodes_to_Variable_Nodes}} \HL{gives the closed-form update~}\HL{\eqref{eq:Messages_from_Factor_Nodes_to_Variable_Nodes_Linear}} \HL{by the Schur complement~}\HL{\cite{zhangSchurComplement2005}}\HL{. Supplementary Material, Sec.~S3, provides the collision-factor Jacobian derivation and a numerical calculation example.}

\section{\HL{Distributed Dynamic Multi-arm Motion Planning Method}} \label{sec:Multi-arm_Motion_Planning_Framework} 

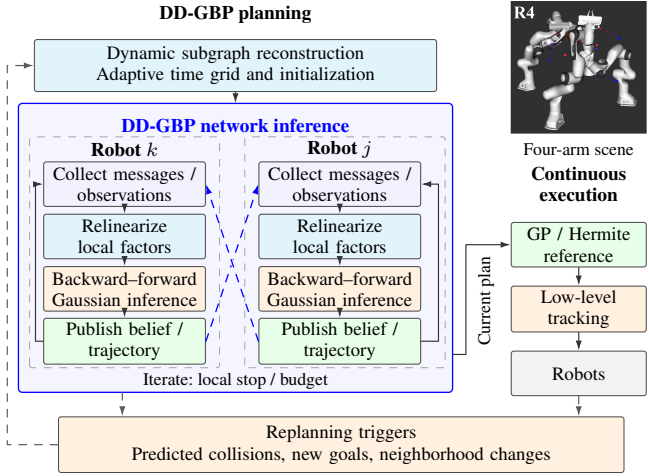
\begin{figure}
\centering
\resizebox{\columnwidth}{!}{%
\input{figures/ddgbp_framework_compact.tex}%
}
\caption{\HL{DD-GBP planning and execution, with an illustrative four-arm scene at the upper right. Each local Gaussian update uses one backward--forward sweep with fixed linearizations and received messages. Robot-wise iterations stop at a local criterion or budget. Blue dashed links exchange messages; gray dashed feedback triggers reconstruction. GP interpolation provides continuous references.}}
\label{fig:Framework}
\end{figure}

\HL{This section presents the planning problem and the operations of DD-GBP} in \HL{the order shown} in Fig.~\ref{fig:Framework}. \HL{We first formulate multi-arm motion planning as inference over GP trajectory parameters, then define the local and inter-arm factors that form the DD-GBP net. Finally, we describe how robots reconstruct their active subnets, schedule local inference and message exchange, and generate continuous execution references. Section~}\HL{\ref{sec:Method_Analysis}} \HL{analyzes the resulting inference accuracy, switching consistency, and connection between GP planning and execution.}

\HL{Unlike MPC methods that explicitly formulate planning as nonlinear programs~\cite{ferrantiDistributedNonlinearTrajectory2023,schwartingParallelAutonomyAutomated2017}, DD-GBP uses an inference--execution--retriggering loop: GP support states represent the trajectory, weighted factors encode boundary, motion, and collision requirements, and local message exchange refines the trajectory through distributed Gaussian inference. Our inference formulation enables asynchronous coordination and flexible anytime online planning.}

\subsection{\HL{GP-Parameterized Planning as MAP Inference}}
\label{sec:Problem_Formulation}

We consider a multi-arm motion planning problem where multiple \HL{arms} $\{R\}_{k=1}^{K}$ are required to move to their target states $\{\bsx_{k}^{\text{goal}}\}_{k=1}^{K}$ from their initial states $\{\bsx_{k}^{\text{init}}\}_{k=1}^{K}$ while avoiding collisions with each other and the environment obstacles $\{O\}_{m=1}^{M}$. When the goal state is reached, the arm will continue to patrol among multiple target states, depending on the task requirements.
\HL{The ideal task is to find continuous joint-space trajectories} \HL{$\{\bsx_k(t)\}_{k=1}^{K}$} \HL{satisfying these requirements. The finite-dimensional, soft-constraint approximation used by DD-GBP is defined below.}
\HL{For arm} \HL{$R_k$}\HL{, we use support states} \HL{$\{\bsx_k(t_i^k)\}_{i=0}^{N_k-1}$} \HL{at timestamps} \HL{$t_i^k=t_0^k+i\Delta t_k$}\HL{. Both} \HL{$N_k$} \HL{and} \HL{$\Delta t_k$} \HL{are selected at a replanning trigger and held fixed during the local solve. Each state} \HL{$\bsx_k(t_i^k)=[\bsq_k(t_i^k);\dot{\bsq}_k(t_i^k)]\in\mbbR^{N_{\mathrm{dim}}}$} \HL{contains} \HL{$N_q$} \HL{joint angles and velocities, with} \HL{$N_{\mathrm{dim}}=2N_q$}\HL{. These support states parameterize a continuous trajectory under the GP prior in Sec.~}\HL{\ref{sec:Dynamic_Factor}}\HL{. In single-arm formulas below,} \HL{$t_i$} \HL{and} \HL{$\Delta t$} \HL{denote the corresponding support timestamps and current interval.}

\HL{Let} \HL{$X_k=\{\bsx_k(t_i)\}_{i=0}^{N_k-1}$} \HL{denote the support states of arm} \HL{$R_k$}\HL{, and let} \HL{$X=\{X_k\}_{k=1}^{K}$} \HL{collect all arms' support states. Each} \HL{$X_k$} \HL{parameterizes a continuous GP trajectory rather than an independent list of waypoints, as detailed in Sec.~}\HL{\ref{sec:Dynamic_Factor}}\HL{. At each replanning instant, inference over these finite-dimensional parameters minimizes}
\HL{\begin{equation}
\label{eq:Problem_Formulation}
\begin{aligned}
X_t^\star&\in\arg\min_X E_t(X),\\
E_t(X)&=\sum_{k=1}^{K}\!\Big(\Phi^k_{\mathrm{init}}+\Phi^k_{\mathrm{goal}}+\sum_i\Phi^{k,i}_{\mathrm{box}}\\
&\quad+\sum_i\Phi^{k,i}_{\mathrm{obs}}+\sum_i\Phi^{k,i}_{\mathrm{mt}}\Big)
+\sum_{(k,j)\in\mathcal E(t)}\sum_i\Phi^{k,j,i}_{\mathrm{inter}}.
\end{aligned}
\end{equation}}
\HL{In the obstacle-cost sum,} \HL{$i$} \HL{indexes both support-state and selected intermediate-time evaluations; the latter depend on their two adjacent support states through GP interpolation. The factors below encode desired conditions as zero-target residuals. Setting} \HL{$\bsz_j=\mathbf0$} \HL{in~}\HL{\eqref{eq:gaussian_residual_factor}} \HL{gives} \HL{$\Phi_{(\cdot)}=\tfrac12\|\bsh_{(\cdot)}\|^2_{\boldsymbol{\Lambda}_{(\cdot)}}$}\HL{. A zero residual therefore represents satisfaction of the corresponding condition. The interaction edge set} \HL{$\mathcal E(t)$} \HL{changes with the range} \HL{$r_{\mathrm c}$} \HL{and event triggers (Sec.~}\HL{\ref{sec:Multi-arm_GBP_Net}}\HL{). Initial and} goal \HL{states, joint and velocity limits, and collision avoidance enter through finite-weight factors. The motion-transition factor supplies the GP prior, while the factor weights determine the relative emphasis on boundary accuracy, motion regularity, and collision avoidance. Increasing a weight penalizes departures from its associated condition more strongly, allowing the planner to adjust the balance among these requirements.}

\HL{DD-GBP performs distributed Gaussian inference to solve this switching MAP problem through local tree updates and inter-arm message exchange, as detailed in Sec.~}\HL{\ref{sec:Distributed_Dynamic_Planning}}\HL{.}

\subsection{Multi-arm DD-GBP Net}
\label{sec:Multi-arm_GBP_Net}
\begin{figure}
\centering
\includegraphics[width=1\linewidth]{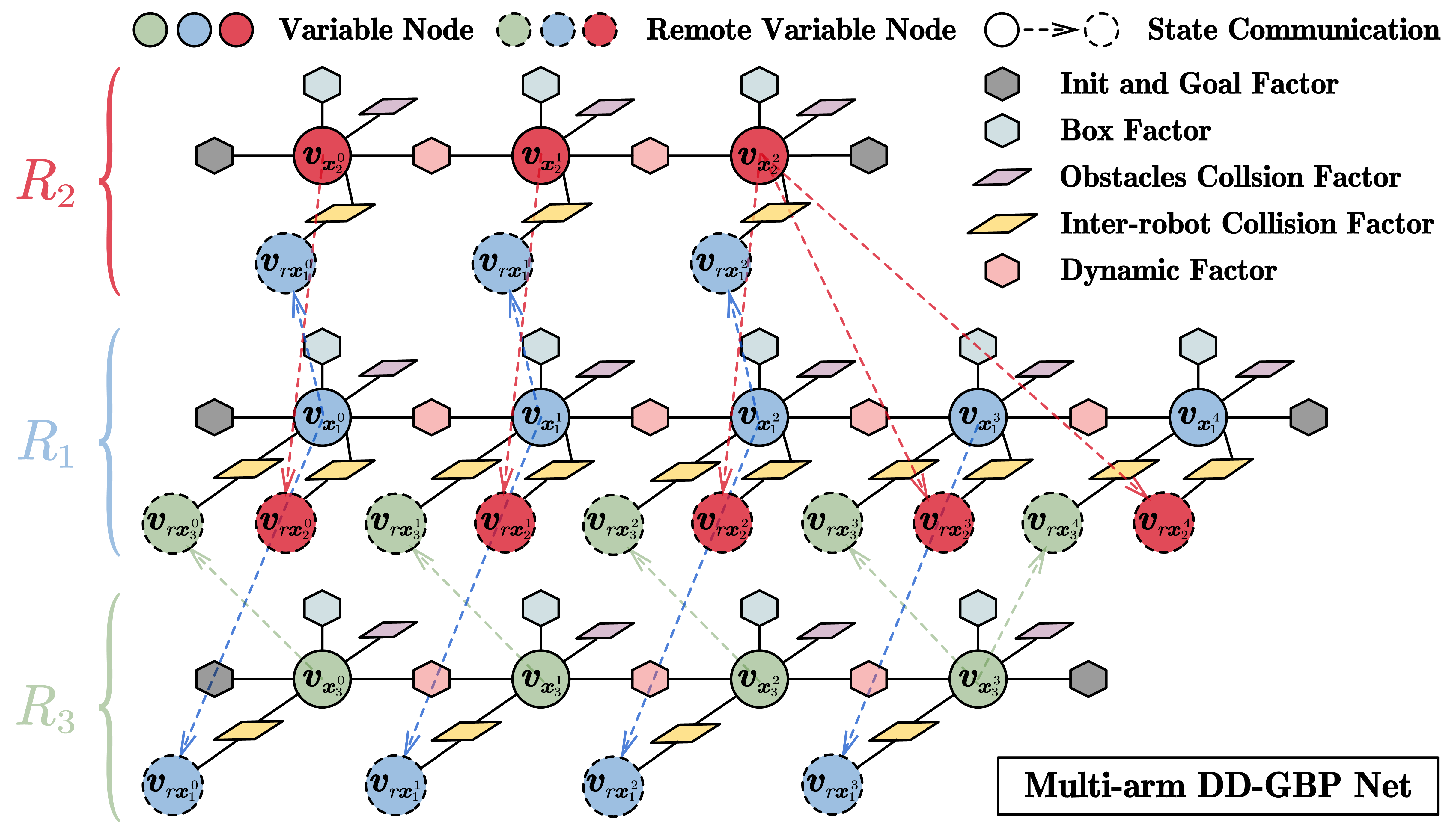}
\caption{Illustration of the proposed multi-arm DD-GBP factor graph network. Solid circles denote each robot's own variable nodes, while dashed circles represent locally stored remote variable nodes from other robots; hexagons and parallelograms denote different types of factor nodes corresponding to motion and interaction constraints. Dashed arrows indicate inter-arm communication links for distributed inference. \HL{For clarity, the schematic omits the interpolated obstacle factors connecting adjacent support states.}}
\label{fig:DD-GBP-Net}
\end{figure}

Fig.~\ref{fig:DD-GBP-Net} \HL{shows one active DD-GBP subnet. For robot} \HL{$R_k$}\HL{, the local graph contains its own trajectory variables} \HL{$\{v_{\bsx_k^i}\}_{i=0}^{N_k-1}$}\HL{, the associated self factors, and remote copies} \HL{$\{v_{\bsx_{\mathrm{r}j}^i}\}_{i=0}^{N_k-1}$} \HL{of neighboring trajectories inside the current interaction range. Static self factors remain attached} to \HL{the trajectory chain, whereas obstacle and inter-arm factors together with the corresponding remote nodes are inserted, removed, or relinearized online as the geometry changes. This trajectory chain and its attached branches determine the local inference schedule described in Sec.~}\HL{\ref{sec:Distributed_Messages_Passing_and_Updating}}\HL{. Dashed arrows indicate the state messages sent from} $R_k$ \HL{to the corresponding remote copies stored} by neighboring \HL{robots.} \HL{The} factor graph network \HL{distinguishes five factor types according to their residuals}. For clarity, the subscript $k$ in $\bsx_k^i$ is omitted when presenting variables and factors in the single-arm local graph, as it remains identical across all \HL{arms} $R_k$.

\subsubsection{Initial and Goal Factor} \label{sec:Initial_and_Goal_Factor} The \HL{boundary residuals} are \HL{$\bsh^{\mathrm{init}}(\bsx^0)=\bsx^0-\bsx^{\mathrm{init}}$} \HL{and} \HL{$\bsh^{\mathrm{goal}}(\bsx^{N_k-1})=\bsx^{N_k-1}-\bsx^{\mathrm{goal}}$}\HL{. Their information matrices are} \HL{$\boldsymbol{\Lambda}^{\mathrm{init}}=\boldsymbol{\Lambda}^{\mathrm{goal}}=\lambda_p\mathbf I_{N_{\mathrm{dim}}}$}\HL{. A} large \HL{$\lambda_p>0$} \HL{gives boundary accuracy a high weight in the objective, keeping the inferred endpoint states close to the prescribed initial and goal states}.

\subsubsection{Box Factor} \label{sec:Box_Factor} Box factor penalizes any violation of joint or velocity limits \HL{through}
\begin{equation}
 \bsh_{\text{box}}(\bsx^i) = \max\left( \bsx^i - \bar{\bsx}, \, \mathbf{0} \right) + \min\left( \bsx^i - \underline{\bsx}, \, \mathbf{0} \right)
\end{equation}
where $\bsx^i = [\boldsymbol{q}^i, \dot{\boldsymbol{q}}^i]$. $\underline{\bsx} = [\underline{\boldsymbol{q}}, \underline{\dot{\boldsymbol{q}}}]$ and $\bar{\bsx} = [\bar{\boldsymbol{q}}, \bar{\dot{\boldsymbol{q}}}]$ are the joint position and velocity limits, and the precision matrix is $\boldsymbol{\Lambda}_{\text{box}} = \lambda_{b}\mathbf{I}_{N_{\text{dim}}}$, where $\lambda_{b} > 0$.

\subsubsection{\HL{Obstacle Collision Factor}} \label{sec:Collision_Factor} \HL{Obstacle and inter-robot collision factors share the hinge residual}
\HL{\begin{equation}
\bsh_c=[1-d_c/\epsilon_c]_+,\qquad c\in\{\mathrm{obs},\mathrm{inter}\},
\label{eq:collision_residual}
\end{equation}}
\HL{where} \HL{$[a]_+=\max(a,0)$}\HL{. For obstacles,} \HL{$d_{\mathrm{obs}}=d_{\mathrm{obs}}(\bsx^i,\{O_m\}_{m=1}^{M})$} \HL{is the minimum signed arm--obstacle distance.} The \HL{safety margin} \HL{$\epsilon_{\mathrm{obs}}>0$} \HL{sets the penalty and trigger threshold, with precision} \HL{$\boldsymbol{\Lambda}_{\mathrm{obs}}=\lambda_o\mathbf I_{1\times1}$}\HL{,} \HL{$\lambda_o>0$}\HL{. For one obstacle,}
\HL{\begin{equation}
d_{\text{obs}}(\bsx^i,O_m)=
\begin{cases}
\min\limits_{p \in \partial R_k,\, q \in \partial O_m}\left\|p(\bsx^i)-q\right\|_2,
& p \notin O_m, \\
-\min\limits_{p \in \partial R_k,\, q \in \partial O_m}\left\|p(\bsx^i)-q\right\|_2,
& p \in O_m,
\end{cases}
\label{eq:Signed_Distance}
\end{equation}}
\HL{where the symbol} \HL{$\partial$} \HL{denotes the spherical set's boundary in the space the object covers, as illustrated in Fig.~\ref{fig:collision_model}(a); if a collision occurs, the signed obstacle distance is negative. We encapsulate each joint of the robot in spheres, which permits efficient distance and gradient computation. The overall obstacle distance is} $d_{\text{obs}}(\bsx^i, \{O_m\}_{m=1}^{M}){=}\min_{m} (d_{\text{obs}}(\bsx^i,O_m))$.

\subsubsection{Inter-robot Collision Factor} \label{sec:Inter-robot_Collision_Factor} 
\begin{figure}[t]
\centering
\begin{minipage}[b]{0.55\linewidth}
\centering
\includegraphics[width=\linewidth]{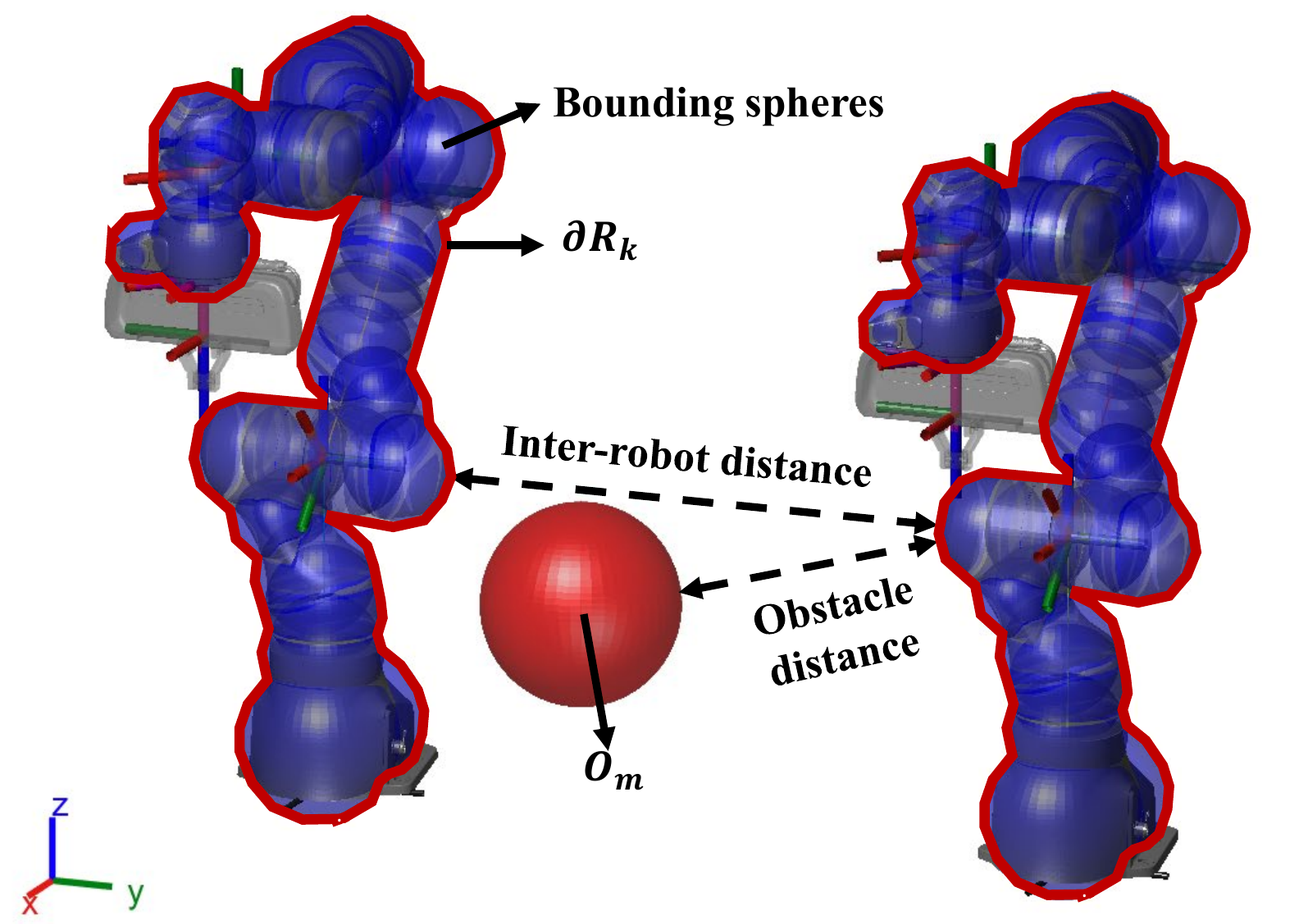}\par
\footnotesize \HL{(a) Collision geometry}
\end{minipage}\hfill
\begin{minipage}[b]{0.43\linewidth}
\centering
\includegraphics[width=\linewidth]{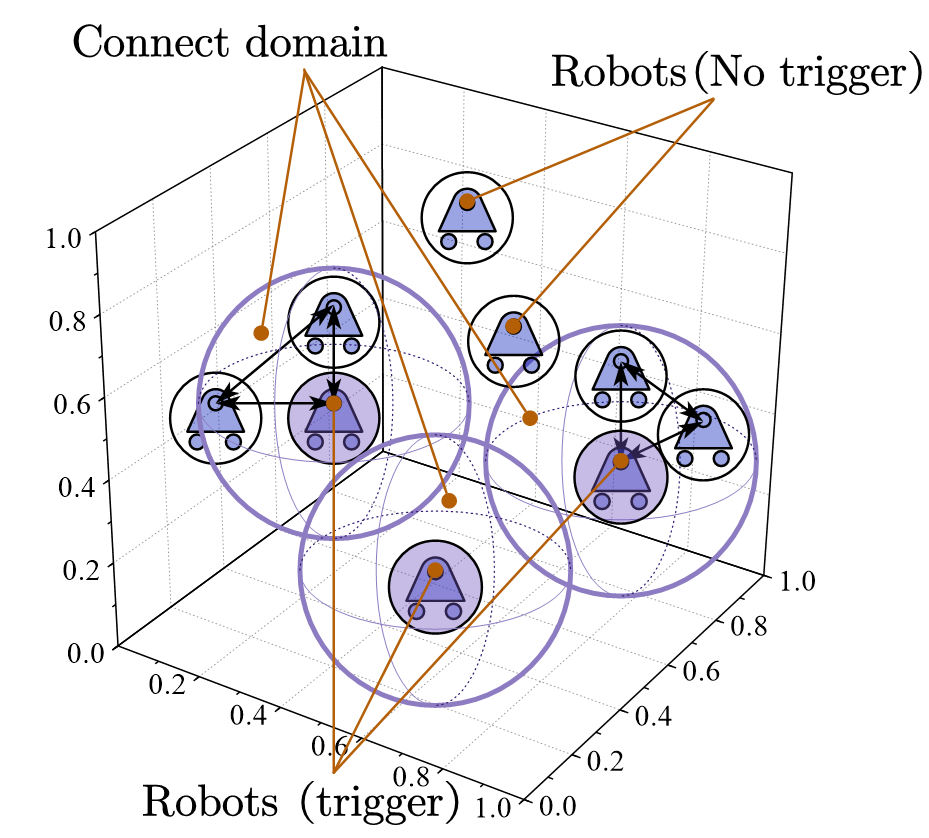}\par
\footnotesize \HL{(b) Triggered neighborhoods}
\end{minipage}
\par\vspace{2pt}
\begin{minipage}[b]{0.55\linewidth}
\centering
\includegraphics[height=18mm,trim=19.20bp 0bp 0bp 348.65bp,clip]{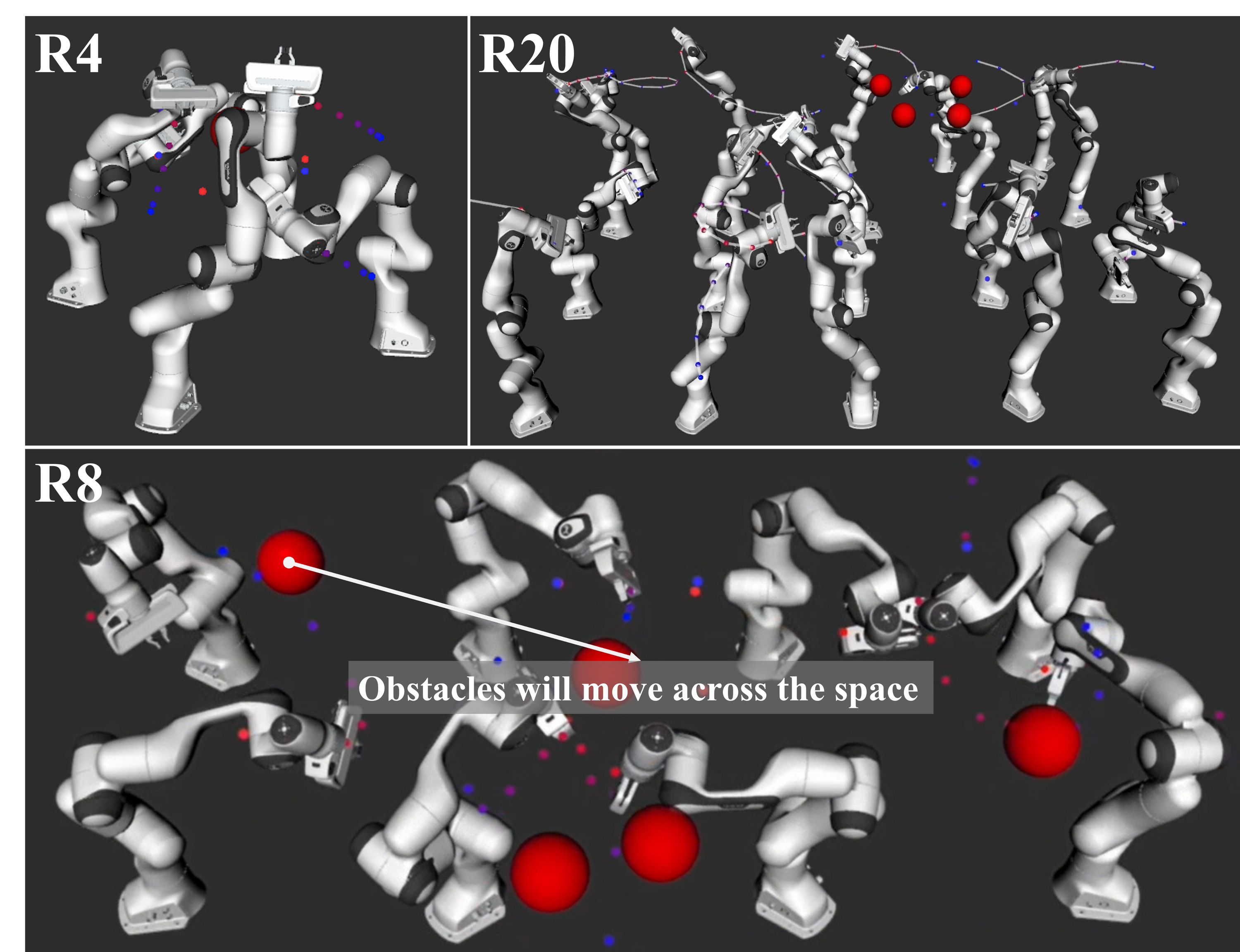}
\end{minipage}\hfill
\begin{minipage}[b]{0.43\linewidth}
\centering
\includegraphics[height=18mm,trim=364.36bp 392.29bp 0bp 12.65bp,clip]{figures/optimized/robot_layout_1_q97.jpg}
\end{minipage}\par
{\footnotesize\HL{(c) Example layouts: 8 arms (left), 20 arms (right)}\par}
\caption{\HL{(a)} Robots are encapsulated by bounding spheres, with robot-obstacle and inter-robot distances defined by the minimum separation between their boundaries. \HL{(b) Triggered robots and their local interaction neighborhoods. (c) Illustrative multi-arm simulation scenes; scalability tests omit the external obstacles shown here.}}
\label{fig:collision_model}
\end{figure}

\HL{In~}\HL{\eqref{eq:collision_residual}}\HL{,} \HL{$d_{\mathrm{inter}}=d_{\mathrm{inter}}(\bsx_k^i,\bsx_j^i)$} \HL{is the minimum signed} distance between \HL{two} arms, \HL{computed} as \HL{in}~\eqref{eq:Signed_Distance}\HL{. Its margin is} \HL{$\epsilon_{\mathrm{inter}}>0$}\HL{, with precision} \HL{$\boldsymbol{\Lambda}_{\mathrm{inter}}=\lambda_i\mathbf I_{1\times1}$}\HL{,} \HL{$\lambda_i>0$}\HL{.} Fig.~\ref{fig:collision_model}\HL{(a)} \HL{shows the bounding-sphere model, while Fig}.\HL{~}\HL{\ref{fig:collision_model}}\HL{(b) illustrates the local interaction neighborhoods used for triggered replanning.}

\subsubsection{\HL{Motion Transition Factor}} \label{sec:Dynamic_Factor}
The \HL{motion-transition factor is a linear Gaussian prior linking consecutive position--velocity support states. It supplies the continuous GP trajectory model; online topology switching supplies the} dynamic \HL{interaction structure. The prior is induced by the linear time-varying stochastic differential equation}
\HL{\begin{equation}\label{eq:gp_prior_sde}
\dot{\bsx}(t)=A(t)\bsx(t)+F(t)\mathbf{w}(t),
\qquad
\mathbf{w}(t)\sim\mathcal{GP}(0,Q_d\,\delta(t-t')),
\end{equation}}
\HL{where} \HL{$\mathbf w(t)$} \HL{is a zero-mean Gaussian white-noise process with spectral density} \HL{$Q_d$}\HL{,} \HL{$\delta(t-t')$} \HL{denotes the Dirac delta, and the noise enters only through the acceleration channel. In the implementation used here,} \HL{$A(t)$} \HL{and} \HL{$F(t)$} \HL{reduce on each interval to the constant-velocity model, so the GP prior between two consecutive support states is represented exactly by the motion transition} factor
\HL{$\bsh_{\text{d}}(\bsx^{i},\bsx^{i+1})=\boldsymbol{\Phi}(t_{i+1},t_i)\bsx^i-\bsx^{i+1}$}\HL{,}
\HL{where} \HL{$\boldsymbol{\Phi}(t_{i+1}, t_{i}) = \begin{bmatrix}
 \mathbf{I}_{N_{\text{q}}} & \Delta t \mathbf{I}_{N_{\text{q}}} \\
 \mathbf{0}_{N_{\text{q}}} & \mathbf{I}_{N_{\text{q}}}   \\
      \end{bmatrix}$}\HL{,} \HL{$\Delta t = t_{i+1}{-}t_{i}$} is the time interval between two consecutive waypoints and \HL{$\boldsymbol{\Lambda}_{\text{d}}=\begin{bmatrix}
 \frac{1}{3}\Delta t^3 Q_d & \frac{1}{2}\Delta t^2 Q_d \\
 \frac{1}{2}\Delta t^2 Q_d & \Delta t Q_d
     \end{bmatrix}^{-1}$} \HL{and} \HL{$Q_d=\lambda_d^{-1}\mathbf{I}_{N_{\text{q}}},\ \lambda_d>0$}.
\HL{The precision weight $\lambda_d$ controls motion regularity relative to the boundary and collision factors: larger precision favors smaller velocity changes, while lower precision allows greater trajectory flexibility. DD-GBP adapts this weight to the time grid, as described in Sec.~\ref{sec:Dynamic_Reconstruction}.}

\HL{Together, the support states and motion prior define a continuous GP trajectory, allowing collision factors to be evaluated at selected intermediate times} \HL{$t_i<\tau<t_{i+1}$} \HL{without adding independent trajectory variables. Following~}\HL{\cite{mukadamContinuoustimeGaussianProcess2018}}\HL{, the GP conditional mean is} \HL{$\bsx(\tau)=\mathbf L_i(\tau)\bsx^i+\mathbf L_{i+1}(\tau)\bsx^{i+1}$}\HL{, where the interpolation matrices are determined by the prior and timestamps. Substituting this state into the obstacle residual in Eq.~}\HL{\eqref{eq:collision_residual}} \HL{gives a binary factor connecting the two adjacent support states. These intermediate states are evaluated analytically through local matrix operations. The same GP map supplies remote-state queries and execution references, as described in Sec.~}\HL{\ref{sec:Online_Hierarchical_Planning}} \HL{and Prop.~}\HL{\ref{prop:gp_execution}}\HL{.}

\subsection{\HL{Distributed and Dynamic Planning with GBP}}\label{sec:Distributed_Dynamic_Planning}
\HL{DD-GBP operates as an event-triggered receding-horizon loop. A new target, a predicted local collision, or a neighborhood change triggers reconstruction of the affected subnet} \HL{$\mathcal G_{\mathrm{tr}}$}\HL{. Each robot linearizes its active} factors \HL{and performs backward--forward inference with the current remote messages held fixed during the sweep. The resulting support-state mean is passed} to \HL{GP-consistent Hermite interpolation and the low-level tracking controller. Robots exchange updated messages only along active inter-arm couplings. The outer loop selects the active subnet, while repeated} local \HL{sweeps and message exchanges refine} its trajectory \HL{estimate.} Alg.~\ref{alg:GBP} \HL{summarizes the procedure}.

\begin{algorithm}[t]
\small
\caption{Multi-Arm Planning with DD-GBP}
\label{alg:GBP}
\begin{algorithmic}[1]
\REQUIRE Initial state $\{\bsx_k^0\}_{k=1}^{K}$, goal states $\{\bsx_{k}^{\text{goal}}\}_{k=1}^{K}$, interaction range $r_{\text{c}}$, max iteration $N_{\text{iter}}^{\text{max}}$, $\texttt{triggered} \gets \texttt{true}$.
\WHILE {$\texttt{triggered}$}
\STATE $\texttt{triggered} \gets \texttt{false}$.
\STATE \HL{Identify trigger sources} \HL{$\mathcal{R}_{\mathrm{tr}}$} \HL{(all robots initially); use} Alg.~\ref{alg:trigger} to \HL{obtain} \HL{affected} \HL{robots} \HL{$\mathcal S$} and \HL{subnet} \HL{$\mathcal{G}_{\mathrm{tr}}$}. 
\FORALL {\HL{$R_k \in \mathcal S$} in parallel}
\STATE $N_{\text{iter}} \gets 0$, $\Delta_{E_k} \gets \infty$.
\WHILE{$\Delta_{E_k} \geq \epsilon_{\text{E}}$ \AND $N_{\text{iter}} \leq N_{\text{iter}}^{\text{max}}$}
 \STATE Using Alg.~\ref{alg:BackwardForward} to pass message and update $\mathcal{G}_k$, get new energy $E_k^{\text{new}}$ and beliefs $\{b_k^{\text{new}}(\bsx_k^i)\}_{i=0}^{N_k-1}$.
 \STATE $\Delta_{E_k} \gets |E_k^{\text{new}} - E_k^{\text{old}}|$, $E_k^{\text{old}} \gets E_k^{\text{new}}$, $N_{\text{iter}} \gets N_{\text{iter}} + 1$
 \STATE Exchange messages of remote variable nodes with connected \HL{arms}.
\ENDWHILE
\STATE Extract $\{\bsx_k(t_i)\}_{i=0}^{N_k-1} \gets \{b_k^{\text{new}}(\bsx_k^i)\}_{i=0}^{N_k-1}$ and interpolate to get continuous trajectory $\bsx_k(t)$.
\STATE Execute \HL{the GP-interpolated reference} with \HL{the low-level tracking} controller \HL{(}Sec.~\ref{sec:Online_Hierarchical_Planning}\HL{).}
\IF{\HL{$\exists i\in\{0,\ldots,N_{\text{h}}-1\}: d(\bsq_k^i) < \epsilon_{\text{obs}}$} \OR new target received \OR new neighbors detected}
 \STATE $\texttt{triggered} \gets \texttt{true}$.
\ENDIF
\ENDFOR
\ENDWHILE
\end{algorithmic}
\end{algorithm}

\subsubsection{Dynamic Reconstruction and Replanning Mechanism}\label{sec:Dynamic_Reconstruction}
\noindent\textbf{\HL{Event-triggered replanning.}}\enspace
\HL{Initially, all robots are trigger sources. Thereafter, a robot becomes a trigger source when it receives a new target, predicts a near-future collision over its first $N_{\text{h}}$ intervals, or detects a new neighbor} in \HL{its adjacency list. Each trigger starts a new round of subnet extraction} \HL{and} \HL{reconstruction, avoiding graph rebuilding when} \HL{nothing locally relevant has changed}.

\noindent\textbf{\HL{Localized subnet extraction.}}\enspace
\HL{At the graph level, each trigger rebuilds only the active interaction neighborhood illustrated in Fig.~}\HL{\ref{fig:collision_model}}\HL{(b). Each robot selects up} to \HL{$N_{n_k}$} \HL{nearby neighbors within} $r_{\text{c}}$\HL{, then runs} a bounded breadth-first search (BFS) with \HL{maximum} depth $d_{\max}$ and \HL{activated-count cap} $N_{\text{act}}^{\max}$ from the trigger source to \HL{extract} the affected \HL{subnet} $\mathcal{S}_{k}$. \HL{The resulting replanning scope therefore depends on the active neighborhood rather than the full team size. When the BFS subnets initiated by neighbouring triggers overlap, every robot is assigned to at most one active subnet via the smallest trigger-source identifier rule, so} that each \HL{variable is updated by a unique scheduler within one} replanning \HL{cycle and message duplication is avoided. The per-robot cost of subnet extraction is} \HL{$\mathcal{O}(N_{\mathrm{act}}^{\max}\,N_{n_k})$}\HL{, independent of the team size} \HL{$K$}\HL{, which helps enable the observed scalability up to} \HL{$K{=}30$} \HL{arms reported in Sec.~}\HL{\ref{sec:Simulations_and_Experiments}}\HL{.}

\HL{The distance threshold removes interactions that have no collision penalty, while the neighbor and BFS caps bound the amount of work assigned to a triggered update. Section~}\HL{\ref{sec:Switching_Consistency}} \HL{distinguishes these two roles by establishing when threshold pruning preserves the complete objective and quantifying the trajectory change induced by a new local planning problem. In the layouts considered here, the spatial extent and packing of the arms limit the number of nearby interacting neighbors, motivating the caps as bounds on local work. The pruning result applies} \HL{when} \HL{these caps retain all contributing collision factors}.

\begin{algorithm}[th]
\small
\caption{Triggered Replanning Subnet Extraction}
\label{alg:trigger}
\begin{algorithmic}[1]
\REQUIRE Triggered robots $\mathcal{R}_{\mathrm{tr}}$, Graph Net $\mathcal{G}$
\ENSURE \HL{Affected robots} \HL{$\mathcal S$} \HL{and rebuilt} subnet $\mathcal{G}_{\mathrm{tr}}$
\STATE Initialize edge sets $\mathcal{E}_k \gets \emptyset$ for all $R_k \in \mathcal{R}_{\mathrm{tr}}$
\FORALL{robot $R_k \in \mathcal{R}_{\mathrm{tr}}$}
\STATE Neighbor set $\mathcal{R}_{n_k} = \{ R_j \mid d_{\text{inter}}(\boldsymbol{x}_k, \boldsymbol{x}_j) < r_{\text{c}} \}$
\STATE Get top $N_{n_k}$ closest robots: $\mathcal{R}_{n_k} \gets \text{sort}(\mathcal{R}_{n_k})[:N_{n_k}]$
\STATE $\mathcal{E}_k \gets \mathcal{E}_k \cup (R_k,R_j)$, $\mathcal{E}_j \gets \mathcal{E}_j \cup (R_j,R_k)$ for all $R_j \in \mathcal{R}_{n_k}$
\ENDFOR
\FORALL{robot $R_k \in \mathcal{R}_{\mathrm{tr}}$}
\STATE Initialize $\mathcal{S}_k \gets \{ R_k \}$, queue $\mathcal{Q}_k \gets \{ R_k \}$, search depth $d_k \gets 0$, activated number $N_{\text{act}} \gets 1$
\WHILE{$\mathcal{Q}_k \neq \emptyset$ and $d_k \leq d_{\max}$}
 \STATE Pop $R_i$ from $\mathcal{Q}_k$
 \FORALL{$(R_i,R_j) \in \text{sort}(\mathcal{E}_i)$ and $N_{\text{act}} \leq N_{\text{act}}^{\max}$}
  \STATE $\mathcal{S}_k \gets \mathcal{S}_k \cup \{R_j\}$, $\mathcal{Q}_k \gets \mathcal{Q}_k \cup \{R_j\}$, $N_{\text{act}} \gets N_{\text{act}} + 1$
 \ENDFOR
 \STATE $d_k \gets d_k + 1$
\ENDWHILE
\ENDFOR
\STATE \HL{Set} \HL{$\mathcal{S} = \bigcup_{R_k \in \mathcal{R}_{\mathrm{tr}}} \mathcal{S}_k$}\HL{; for each} \HL{$R_k\in\mathcal S$}\HL{, select} \HL{$(N_k,\Delta t_k)$} \HL{by Eq.~}\HL{\eqref{eq:Num_Waypoints}}\HL{.}
\STATE Construct \HL{$\mathcal{G}_{\mathrm{tr}} \gets \mathcal{G}[\mathcal{S}]$} \HL{on the new timestamps;} initialize \HL{support states and update} the \HL{GP} \HL{and} \HL{collision} \HL{factors.}
\end{algorithmic}
\end{algorithm}

\noindent\textbf{\HL{Adaptive planning horizon.}}\enspace
\HL{For each robot in the extracted subnet, we} \HL{adapt the planning horizon to the remaining motion} \HL{by jointly selecting the support count} \HL{$N_k$} \HL{and interval} \HL{$\Delta t_k$}\HL{. Let} \HL{$\bar v_{k,a}>0$} \HL{be the velocity limit of joint} \HL{$a$}\HL{,} \HL{with} \HL{$N_{\max}\ge N_{\min}\ge3$} \HL{and} \HL{$\Delta t_{\max}\ge\Delta t_{\min}>0$}\HL{. We estimate the remaining motion time} \HL{$T_k=\max_a|q^{\mathrm{goal}}_{k,a}-q^0_{k,a}|/\bar v_{k,a}$} \HL{from the joint displacements} \HL{and velocity limits}, \HL{then set}
\HL{\begin{equation}
\begin{aligned}
\rho_k&=\operatorname{clip}_{[0,1]}\!\left(\frac{T_k}{(N_{\max}-1)\bar\Delta_k}\right),\\
N_k&=\operatorname{round}\!\left(N_{\min}+\rho_k(N_{\max}-N_{\min})\right),\\
\Delta t_k&=\operatorname{clip}_{[\Delta t_{\min},\bar\Delta_k]}\!\left(\frac{T_k}{N_k-1}\right).
\end{aligned}
\label{eq:Num_Waypoints}
\end{equation}}
\HL{where} \HL{$\operatorname{clip}_{[a,b]}(z)=\min\{b,\max(a,z)\}$} \HL{and} \HL{$\operatorname{round}$} \HL{selects} the \HL{nearest} \HL{integer.} \HL{The joint-adaptation defaults are} \HL{$N_{\min}=3$}, \HL{$\Delta t_{\min}=0.01\,\mathrm s$}\HL{, and} \HL{$\Delta t_{\max}=0.6\,\mathrm s$}\HL{. Normally,} \HL{$\bar\Delta_k=\Delta t_{\max}$}\HL{. If a near-obstacle cap is enabled} and the \HL{current obstacle} \HL{clearance} \HL{$c_k<d_{\mathrm{cap}}$}\HL{, we first set} \HL{$\bar\Delta_k=\max\{\Delta t_{\min},\min(\Delta t_{\max},\Delta t_{\mathrm{obs}})\}$} \HL{and use this bound throughout} Eq.~\eqref{eq:Num_Waypoints}\HL{.}

\HL{Far from the goal, more support states and larger intervals cover a longer motion. As the goal approaches, fewer states are needed and their spacing is reassigned from the remaining time; after} \HL{$N_k$} \HL{reaches} \HL{$N_{\min}$}\HL{, the interval can continue to decrease to} \HL{$\Delta t_{\min}$}\HL{. Retaining at least three states leaves an interior state available for inference between the start and goal, avoiding a single-interval trajectory. The resulting horizon is} \HL{$H_k=(N_k-1)\Delta t_k$}\HL{; when the remaining time exceeds its allowed span, receding-horizon updates continue progress toward the goal.}

\noindent\textbf{\HL{Adaptive GP scaling.}}\enspace
\HL{Let $\bar a$ be the prescribed acceleration bound. Omitting the robot index as above, we set}
\HL{\begin{equation}
\lambda_d(\Delta t)=\frac{\sigma_z^2}{\bar a^2\Delta t},\qquad
Q_d(\Delta t)=\lambda_d(\Delta t)^{-1}\mathbf I.
\label{eq:adaptive_process_scaling}
\end{equation}}
\HL{For the prior velocity increment $\Delta v$, each joint's interval-averaged acceleration disturbance $a_{\mathrm{av}}=\Delta v/\Delta t$ has standard deviation $1/\sqrt{\lambda_d\Delta t}=\bar a/\sigma_z$. With $\sigma_z=3$, the prior assigns approximately $99.73\%$ probability to $|a_{\mathrm{av}}|\le\bar a$ per joint and interval. A larger allowed $\bar a$ lowers the norm of $\boldsymbol\Lambda_{\mathrm d}$ and softens the trajectory, allowing stronger responses to obstacles. Excessive softness can amplify oscillations or lengthen execution. Adapting $\lambda_d$ therefore balances responsiveness and trajectory sensitivity at the prescribed acceleration scale, updating the motion covariance and precision before collision evaluation on the new grid.}

\noindent\textbf{\HL{Initialization and graph update.}}\enspace
\HL{For a newly assigned goal, the trajectory is initialized with} a \HL{straight-line} \HL{joint-space} \HL{seed, whereas replanning toward an unchanged goal} \HL{is} \HL{warm-started by evaluating the} previous \HL{GP trajectory at the new support timestamps using the same interpolation map as online execution. The self-factor chain is resized to the selected grid; obstacle/ inter-arm factors and the corresponding remote variable nodes are inserted, removed, or relinearized as the local geometry changes. Neighbor states are aligned by timestamp on this grid, providing the remote inputs for the message-passing procedure below.}

\subsubsection{Distributed \HL{Message} Passing and Updating} \label{sec:Distributed_Messages_Passing_and_Updating}

\begin{figure}
\centering
\includegraphics[width=0.9\linewidth]{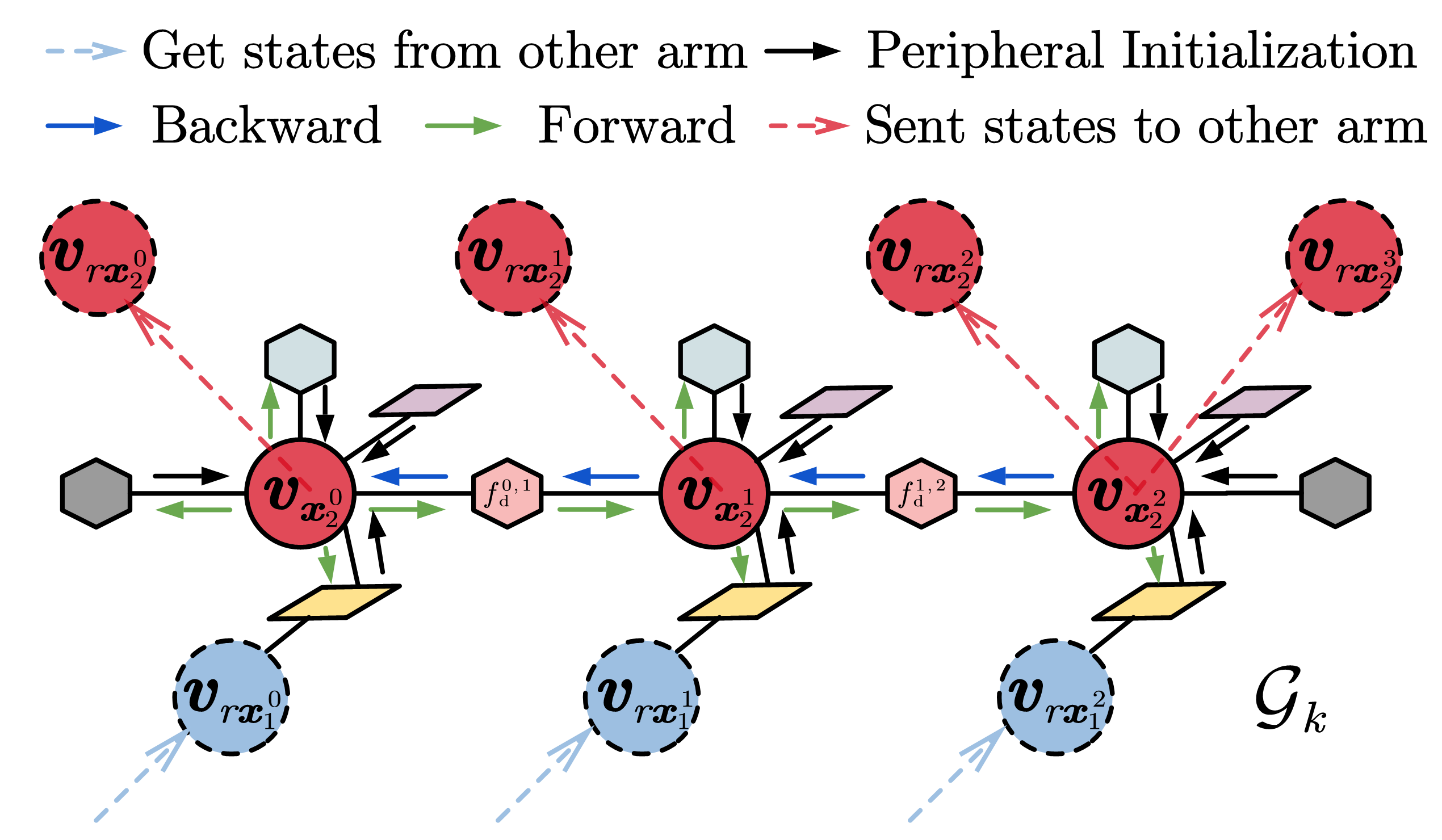}
\caption{Backward–forward message passing for each arm's local tree-structured graph. The process consists of a peripheral initialization step (black arrows), followed by a backward pass from the goal to the start (blue arrows), and a forward pass from the start to the goal (green arrows), and state exchanges at the beginning and end (dashed arrows).}
\label{fig:BackwardForward}
\end{figure}

In DD-GBP, \HL{message passing and belief updates dominate the online cost. We propose a tree-directed local schedule that replaces the synchronous updates of our Syn-GBP planner~\cite{jin2026distributedSafetyAware}. Fig.~\ref{fig:BackwardForward} and Alg.~\ref{alg:BackwardForward} show the procedure.}

\HL{For robot $R_k$, the active subgraph $\mathcal G_k$ contains its trajectory chain, boundary, box, obstacle, and inter-arm factors, with neighbors represented by remote variable nodes. Each backward--forward sweep fixes the factor linearizations and received neighbor messages, propagates information from the goal and then from the start, and recovers the local Gaussian marginal means and covariances. Updated messages are exchanged only along active inter-arm links; relinearization and subsequent exchanges refine the coupled trajectories. Section~}\HL{\ref{sec:Convergence_and_Optimality_of_GBP}} \HL{relates this block schedule to convergence of the coupled graph.}

\begin{remark}
 In the state exchange step, each \HL{arm} $R_k$ may have a different \HL{support count} $N_k$ \HL{and interval} \HL{$\Delta t_k$}. \HL{Since each local plan is a parameterized GP trajectory rather than a fixed index sequence, message} exchange \HL{is aligned by timestamp. When} \HL{$R_k$} \HL{needs the state of a neighbor} \HL{$R_j$} \HL{at its own timestamp} \HL{$t_i^k$}\HL{, it queries or interpolates} \HL{$\bsx_j(t_i^k)$} \HL{from} \HL{$R_j$}\HL{'s current GP support states over the shared time window. Therefore different} \HL{$N_k\neq N_j$} \HL{or} \HL{$\Delta t_k\neq\Delta t_j$} \HL{do not create ambiguity; robots exchange only the beliefs needed to reconstruct time-aligned remote states. Beyond a neighbor's available horizon, its last available joint position is held with zero velocity, corresponding to} the \HL{assumption} \HL{that it remains stationary after reaching its terminal state. For each} \HL{arm}\HL{, subgraph updates continue whenever new} remote messages \HL{arrive} and repeat until the local state remains stable after incorporating all updates or the maximum iteration is reached.
\end{remark}

\begin{algorithm}[htbp]
\small
\caption{Backward–Forward Message Passing for $\mathcal{G}_k$}
\label{alg:BackwardForward}
\begin{algorithmic}[1]
 \REQUIRE Previous beliefs $\{b_{\text{old}}(\bsx_k^i)\}$ and messages, local graph $\mathcal{G}_k$
 \ENSURE Updated beliefs $\{b_{\text{new}}(\bsx_k^i)\}$ and energy $E_k^{\text{new}}$
 \STATE $b_{\text{new}}(\bsx_k^i) \gets b_{\text{old}}(\bsx_k^i)$, warm start $m_{w \to v}, m_{v \to w}$
 \STATE Collect neighbor \HL{messages and obstacle observations; linearize local factors via} \HL{\eqref{eq:Factor_Linearization}} and compute \HL{peripheral} messages\HL{. Keep factors and remote messages fixed for this sweep}.
 \STATE \textbf{Backward pass:}
 \FOR{$i = N_k - 1$ down to $1$}
 \STATE Update variable belief $b_{\text{new}}(\bsx_k^i)$ \HL{as in Sec}.~\HL{\ref{sec:Variable_Belief_Update}}
 \STATE Pass message $\bsx_k^i \to f^{i-1,i}_{\text{d}}$, $f^{i-1,i}_{\text{d}} \to \bsx_{i-1}$ via 
 Eq.~\eqref{eq:Messages_from_Variable_Nodes_to_Factor_Nodes}-\eqref{eq:Messages_from_Factor_Nodes_to_Variable_Nodes}
 \ENDFOR
 \STATE \textbf{Forward pass:}
 \FOR{$i = 0$ to $N_k - 1$} 
 \STATE Update $b_{\text{new}}(\bsx_k^i)$ with all incoming messages 
 \STATE Pass messages to all neighbor factors except $f^{i-1,i}_{\text{d}}$ if exists.
 \STATE \HL{Using the fixed linearization}, pass \HL{$f^{i,i+1}_{\mathrm d}$} to $\bsx_k^{i+1}$ if \HL{it} exists.
 \ENDFOR
 \RETURN $\{b_{\text{new}}(\bsx_k^i)\}$ and compute the total factor energy $E_k^{\text{new}}$. 
\end{algorithmic}
\end{algorithm}

\subsubsection{\HL{GP-Consistent Interpolation and Online Execution}}
\label{sec:Online_Hierarchical_Planning}
\HL{After a local update, the robot uses the inferred position--velocity support-state means to construct a continuous reference through Eq.~}\HL{\eqref{eq:gp_hermite_execution}}\HL{. The reference} \HL{$\bsx_r(t)=[\bsq_r(t);\dot{\bsq}_r(t)]$} \HL{passes through a joint-velocity saturation limiter before being supplied to the low-level tracking controller. For the adopted constant-velocity prior driven by white acceleration, the nominal cubic Hermite reference before saturation is the GP conditional mean, as established in Prop.~}\HL{\ref{prop:gp_execution}}\HL{.}

\HL{Intermediate-time collision evaluation, timestamp-aligned remote-state queries, and warm-start transfer use the same interpolation map. With remote messages or trajectories frozen during a local sweep, each local interpolation factor acts on its adjacent support pair; factors on that pair are combined with the motion-transition factor to retain the local tree structure. New targets, predicted near-future collisions, and neighborhood changes trigger updates of the affected subnet. If the prescribed iteration budget is reached before convergence, the current mean remains an executable anytime trajectory; the time-aligned warm start then carries it into subsequent cycles for continued refinement. Section~}\HL{\ref{sec:Method_Analysis}} \HL{states the conditions for exact local inference and the bounds used to assess an early-terminated update.}

\section{\HL{Inference Accuracy and Trajectory Consistency of DD-GBP}}
\label{sec:Method_Analysis}
\HL{This section examines how the DD-GBP design supports accurate local updates and consistent trajectory execution. We first analyze the Gaussian chain solve and its anytime error bounds, then quantify the effect of} \HL{subnet changes on successive plans, and finally establish the common GP representation used for planning and execution.}

\subsection{\HL{Convergence and Anytime Planning}}
\label{sec:Convergence_and_Optimality_of_GBP}
\HL{For nonlinear GBP on loopy graphs, convergence under repeated relinearization and topology changes remains an open problem in general. For the fixed Gaussian models covered by Weiss and Freeman~}\HL{\cite{weissCorrectnessBelief2001}}\HL{, converged belief means recover the Gaussian MAP solution even when the covariance estimates are approximate. Walk-summability and strict diagonal dominance provide sufficient convergence conditions~}\HL{\cite{malioutovWalkSumsBelief2006}}\HL{, while analyses of different factorizations, schedules, damping choices, and variable groupings further clarify how the inference design affects convergence~}\HL{\cite{duConvergenceAnalysisDistributed2018,moallemiConvergenceMinSumConvex2010,suConvergenceConditionsGaussian2015,kamperConvergenceGaBPNodes2019}}\HL{.}

\HL{In dynamic multi-arm planning, we regard a timely, usable trajectory estimate as sometimes more important than waiting for full convergence while obstacles and neighboring arms continue to move. The iterative structure of GBP supports this choice, as each completed local sweep provides an estimate for its current Gaussian subproblem, which remains available if the distributed iteration is interrupted. We carry this estimate into the next planning cycle through the time-aligned GP warm start, allowing refinement to continue as new information arrives.}

\HL{We organize these updates through the robot-wise schedule in Fig.~}\HL{\ref{fig:Framework}}\HL{. Before each local solve, a robot collects updated neighbor beliefs or trajectories and obstacle observations as external information. It then holds this information fixed and linearizes its local factors about the current trajectory. By combining interpolation-time factors on each adjacent support pair with the corresponding motion-transition factor, we retain a chain on which one backward--forward sweep gives exact local Gaussian marginals~}\HL{\cite{bishop2006pattern}}\HL{. The robot then publishes its updated belief or trajectory to its neighbors.}

\HL{These alternating local solves and message exchanges form block-wise Gaussian inference. Multivariate GaBP studies relate convergence to variable grouping~}\HL{\cite{kamperConvergenceGaBPNodes2019}}\HL{, motivating our robot-wise blocks: each local solve groups a coupled trajectory chain, while neighbor messages convey cross-robot influence. Each completed sweep returns the MAP trajectory estimate for that robot's current fixed Gaussian subproblem. When an iteration cap or deadline interrupts the coupled solve, the following proposition bounds the remaining error using the assembled residual and local linearization mismatch.}

\begin{proposition}[Anytime early-stopping bound]\label{prop:anytime_bound}
\HL{Consider one relinearization cycle at instant} \HL{$t$} \HL{with linearized per-cycle energy} \HL{$E_t^{\mathrm{lin}}(X)=\tfrac12 X^{\top}H_tX-g_t^{\top}X$}\HL{, where} \HL{$H_t=\sum_{f}J_f^{\top}\Lambda_fJ_f\succeq mI$} \HL{for a common lower bound} \HL{$m>0$}\HL{, and assume the underlying nonlinear per-cycle energy} \HL{$E_t$} \HL{is} \HL{$m$}\HL{-strongly convex over the current trust region with linearization mismatch} \HL{$\|\nabla E_t(X)-\nabla E_t^{\mathrm{lin}}(X)\|_2\le\delta_t$} \HL{there. Let} \HL{$X^{\tau}$} \HL{be the trajectory after} \HL{$\tau$} \HL{inner GBP sweeps and} \HL{$r^{\tau}=H_tX^{\tau}-g_t$} \HL{the assembled linear-system residual, let} \HL{$X_t^{\star}=H_t^{-1}g_t$} \HL{be the linearized minimizer, and let} \HL{$X_{\mathrm{nl},t}^{\star}$} \HL{be an interior local minimizer of} \HL{$E_t$}\HL{. Assume that this minimizer and} \HL{$X^\tau$} \HL{lie in the same strongly convex region. Then}
\HL{\begin{equation}
\|X^{\tau}-X_t^{\star}\|_2\;\le\;\tfrac{1}{m}\,\|r^{\tau}\|_2,
\label{eq:residual_bound}
\end{equation}}
\HL{and the same iterate satisfies the nonlinear early-stopping bound}
\HL{\begin{equation}
\|X^{\tau}-X_{\mathrm{nl},t}^{\star}\|_2\;\le\;\tfrac{1}{m}\big(\|r^{\tau}\|_2+\delta_t\big).
\label{eq:early_stop_cost_bound}
\end{equation}}
\end{proposition}

\begin{IEEEproof}
\HL{Since} \HL{$X^{\tau}-X_t^{\star}=H_t^{-1}r^{\tau}$} \HL{and} \HL{$\|H_t^{-1}\|_2\le 1/m$}\HL{, Eq.~}\HL{\eqref{eq:residual_bound}} \HL{follows. Strong convexity and} \HL{$\nabla E_t(X_{\mathrm{nl},t}^{\star})=0$} \HL{imply}
\HL{\begin{equation}\label{eq:strong_convex_gradient_bound}
m\,\|X^{\tau}-X_{\mathrm{nl},t}^{\star}\|_2\le\|\nabla E_t(X^{\tau})\|_2.
\end{equation}}
\HL{Writing} \HL{$\nabla E_t(X^{\tau})=r^{\tau}+e_t$}\HL{, where} \HL{$e_t=\nabla E_t(X^{\tau})-\nabla E_t^{\mathrm{lin}}(X^{\tau})$} \HL{satisfies} \HL{$\|e_t\|_2\le\delta_t$}\HL{, bounds the right-hand side by} \HL{$\|r^{\tau}\|_2+\delta_t$}\HL{. Division by} \HL{$m$} \HL{proves Eq.~}\HL{\eqref{eq:early_stop_cost_bound}}\HL{.}
\end{IEEEproof}

\HL{In our DD-GBP construction, boundary anchoring and the GP transition model provide the positive definiteness needed for each local Gaussian solve. During the solve, factor linearizations and positive-semidefinite incoming remote information remain fixed, so the information matrix is assembled from positive semidefinite contributions. With positive-definite process covariances and a positive-definite full-state boundary factor, a zero-cost perturbation must vanish at the anchored state and satisfy} \HL{$\delta\bsx^{i+1}=\boldsymbol\Phi(t_{i+1},t_i)\delta\bsx^i$}\HL{, so it vanishes along the entire chain. The information matrix is therefore positive definite, and additional positive semidefinite factors preserve this property. These modeling choices give each local solve a strongly convex quadratic objective and a unique mean/ MAP solution.}

\HL{In our method, the anchored GP prior provides a source of curvature for the nonlinear model as well. The nonlinear bound applies in a smooth local region where this curvature dominates the residual-weighted second-order terms. We use the time-aligned GP warm start and repeated relinearization to keep successive expansion points close to the current trajectory; with bounded residuals and Lipschitz-continuous Jacobians, small departures from these points make the gradient mismatch} \HL{$\delta_t$} \HL{small. The assembled residual and} \HL{$\delta_t$} \HL{then separate the remaining Gaussian inference error from the local approximation error, giving us a basis for choosing a stopping tolerance and assessing the estimate available at a planning deadline.}

\subsection{\HL{Topology-Switching Consistency under Replanning}}
\label{sec:Switching_Consistency}
\HL{We dynamically extract local subgraphs in DD-GBP to focus computation on the interactions relevant to the current update. This gated topology extraction preserves the complete global gated objective when all contributing active factors, free variables, and fixed boundary information are retained, as established in Prop.~}\HL{\ref{prop:gated_equivalence}}\HL{. We then examine how observations and factor updates affect successive plans. Prop.~}\HL{\ref{prop:replanning_consistency}} \HL{bounds the difference between the updated local quadratic trajectory and the time-aligned previous trajectory. Together, these results explain the objective represented by each reconstructed graph and the trajectory change between successive planning cycles.}

\begin{proposition}[Equivalence of hard-gated threshold-aware subnet switching]
\label{prop:gated_equivalence}
\HL{Let} \HL{$\mathcal E^{\mathrm{full}}=\{(k,j):k<j\}$} \HL{and} \HL{$\Phi_{\mathrm{inter}}^{kj}(X)=\tfrac12\|\bsh_{\mathrm{inter}}^{kj}(X)\|_{\boldsymbol\Lambda_{\mathrm i}}^2$}\HL{. Define} \HL{$g_c(z)=\mathbf1_{\{z\le r_c\}}$}\HL{, with} \HL{$r_c\ge\epsilon_{\mathrm{inter}}$}\HL{, and} \HL{$\mathcal E^{\mathrm{sw}}(X)=\{(k,j)\in\mathcal E^{\mathrm{full}}:d_{kj}(X)\le r_c\}$}\HL{. Suppose subnet extraction retains every factor in this active set, with the same free variables and fixed boundary information. Then the gated full energy and switched energy satisfy}
\HL{\begin{equation}
\begin{aligned}
E_t^{\mathrm{full}}(X)
&=E_t^{\mathrm{local}}(X)+\sum_{(k,j)\in\mathcal E^{\mathrm{full}}}g_c\!\big(d_{kj}(X)\big)\Phi_{\mathrm{inter}}^{kj}(X)\\
&=E_t^{\mathrm{local}}(X)+\sum_{(k,j)\in\mathcal E^{\mathrm{sw}}(X)}\Phi_{\mathrm{inter}}^{kj}(X)=E_t^{\mathrm{sw}}(X).
\end{aligned}
\label{eq:gated_energy_equivalence}
\end{equation}}
\HL{Their gradients and Hessians coincide wherever defined. If the activation distance ensures that each inter-arm loss vanishes for} \HL{$d_{kj}(X)\ge\epsilon_{\mathrm{inter}}$} \HL{and is} \HL{$C^p$}\HL{, switching introduces no additional discontinuity through order} \HL{$p$}\HL{. In regions where omitted residuals and Jacobians are zero, pruning also leaves the linearized Gaussian problem unchanged.}
\end{proposition}

\begin{IEEEproof}
\HL{The gate equals one exactly on} \HL{$\mathcal E^{\mathrm{sw}}(X)$}\HL{, giving the energy identity and hence equality of its derivatives wherever they exist. Under the vanishing-loss condition, the gate changes only where the loss is already zero; multiplication by the gate therefore preserves its} \HL{$C^p$} \HL{regularity. An omitted factor with zero residual and Jacobian contributes zero information to the linearized model.}
\end{IEEEproof}

\HL{In local DD-GBP updates, outside-robot trajectories enter as fixed boundary information, and the neighbor and BFS caps are intended to cover the nearby arms allowed by the physical layout. Prop.~}\HL{\ref{prop:gated_equivalence}} \HL{applies when all contributing active factors are retained. For a capped set} \HL{$\mathcal E^{\mathrm{ret}}\subseteq\mathcal E^{\mathrm{sw}}$}\HL{, the additional approximation is exactly the omitted contribution} \HL{$\sum_{(k,j)\in\mathcal E^{\mathrm{sw}}\setminus\mathcal E^{\mathrm{ret}}}\Phi_{\mathrm{inter}}^{kj}$}\HL{, unless these terms vanish on the comparison domain or remain as fixed boundary factors.}

To \HL{study consistency from one replanning cycle to the next, express the two fixed linearizations in a common time-aligned local coordinate vector} \HL{$X$}\HL{. The previous trajectory supplies the warm start} \HL{$\bar X=\mathcal S_{\tau_{\mathrm{shift}}}(X_t^\star)$} \HL{through the same GP interpolation used for execution, where} \HL{$\tau_{\mathrm{shift}}$} \HL{is the elapsed time between planning origins. When the support count or spacing changes, this transfer evaluates the previous trajectory at the new timestamps and places both models in the same support-state coordinates. The following result then relates the new quadratic solution to the transferred trajectory through its remaining gradient imbalance and the factors modified by the trigger.}

\begin{proposition}[Trajectory consistency across triggered replanning updates]
\label{prop:replanning_consistency}
\HL{Let} \HL{$E^-(X)=\sum_f\Phi_f^-(X)$} \HL{and} \HL{$E^+(X)=\sum_f\Phi_f^+(X)$} \HL{be the pre-update reference and updated local quadratic models, where}
\HL{\begin{equation}\label{eq:replan_factor_quadratic}
\Phi_f^\pm(X)=\tfrac12\|A_f^\pm X-b_f^\pm\|_{\Lambda_f^\pm}^2.
\end{equation}}
\HL{Use the union of their factor indices, assigning zero contributions to absent factors, and let} \HL{$\mathcal F_{\mathrm{chg}}$} \HL{include every added, removed, or modified factor. In particular, a changed linearization, weight, observation, or time alignment counts as a modification. Assume} \HL{$H^+=\sum_f(A_f^+)^\top\Lambda_f^+A_f^+\succeq mI$} \HL{with} \HL{$m>0$}\HL{, and let} \HL{$X_+^\star$} \HL{minimize} \HL{$E^+$}\HL{. Define} \HL{$g^-=\nabla E^-(\bar X)$} \HL{and} \HL{$\Delta g_f=\nabla\Phi_f^+(\bar X)-\nabla\Phi_f^-(\bar X)$}\HL{. Then}
\HL{\begin{equation}
\begin{aligned}
\|X_+^\star-\bar X\|_2
&\le\frac1m\|\nabla E^+(\bar X)\|_2\\
&\le\frac1m\left(\|g^-\|_2+\sum_{f\in\mathcal F_{\mathrm{chg}}}\|\Delta g_f\|_2\right).
\end{aligned}
\label{eq:replan_quadratic_bound}
\end{equation}}
\HL{For a prefix of} \HL{$1\le n\le N$} \HL{support states, the consistency metric satisfies}
\HL{\begin{equation}
\mathcal C_n(t,t+1)=\sum_{\ell=0}^{n-1}\|\bsx_{+,\ell}^\star-\bar{\bsx}_\ell\|_2\le\sqrt n\,\|X_+^\star-\bar X\|_2.
\label{eq:replan_prefix_bound}
\end{equation}}
\end{proposition}

\begin{IEEEproof}
\HL{Since} \HL{$E^+$} \HL{is quadratic, its gradient is affine and satisfies} \HL{$\nabla E^+(\bar X)-\nabla E^+(X_+^\star)=H^+(\bar X-X_+^\star)$}\HL{. The second gradient is zero at the minimizer, so inversion of} \HL{$H^+$} \HL{and} \HL{$\|(H^+)^{-1}\|_2\le1/m$} \HL{give the first inequality in Eq.~}\HL{\eqref{eq:replan_quadratic_bound}}\HL{. To obtain the second, subtract the two factor sums at the same point} \HL{$\bar X$}\HL{. Every unchanged factor cancels, and the definitions of} \HL{$g^-$} \HL{and} \HL{$\Delta g_f$} \HL{give}
\HL{\begin{equation}\label{eq:replan_gradient_decomposition}
\nabla E^+(\bar X)=g^-+\sum_{f\in\mathcal F_{\mathrm{chg}}}\Delta g_f.
\end{equation}}
\HL{Applying the triangle inequality yields the stated bound. For the prefix metric, Cauchy--Schwarz bounds the sum of} \HL{$n$} \HL{state-error norms by} \HL{$\sqrt n$} \HL{times their joint Euclidean norm. This prefix norm is no larger than} \HL{$\|X_+^\star-\bar X\|_2$}\HL{, which proves Eq.~}\HL{\eqref{eq:replan_prefix_bound}}\HL{.}
\end{IEEEproof}

\HL{Prop.~}\HL{\ref{prop:replanning_consistency}} \HL{quantifies trajectory consistency between successive planning cycles. When the warm-start imbalance and changed-factor gradient increments are small, the updated quadratic trajectory remains close to the time-aligned previous plan, and the prefix bound measures the change in the portion used for near-term execution. Reusing the previous GP trajectory and updating the factors affected by a trigger make these two sources of change explicit in DD-GBP. If the returned iterate is} \HL{$\widehat X_+$}\HL{,} its \HL{additional error relative to} \HL{$X_+^\star$} \HL{is bounded by} \HL{$\|\nabla E^+(\widehat X_+)\|_2/m$}\HL{, as in Prop.~}\HL{\ref{prop:anytime_bound}}\HL{. The same} \HL{calculation} \HL{can therefore assess each triggered update using its current model and residual.}

\subsection{\HL{Consistency of GP Interpolation and Continuous Execution}}
\label{sec:GP_Execution_Consistency}
\HL{In DD-GBP, we use the inferred means of position--velocity support states to parameterize a continuous GP trajectory. Its motion-transition prior comes from a second-order SDE driven by white Gaussian acceleration, as described in Sec.~}\HL{\ref{sec:Dynamic_Factor}}\HL{. The Markov structure allows the conditional prior mean within each support interval to be evaluated from its two endpoint states. Prop.~}\HL{\ref{prop:gp_execution}} \HL{shows that this evaluation is linear in those states and equals the cubic Hermite reference used in Sec.~}\HL{\ref{sec:Online_Hierarchical_Planning}}\HL{.} \HL{This gives us a direct and efficient connection from the inferred trajectory to continuous execution.}

\begin{proposition}[GP--Hermite execution consistency]
\label{prop:gp_execution}
\HL{On an interval} \HL{$[t_i,t_{i+1}]$} \HL{of length} \HL{$h>0$}\HL{, let} \HL{$\dot{\bsq}=\boldsymbol v$} \HL{and} \HL{$\dot{\boldsymbol v}=\mathbf w$}\HL{, where} \HL{$\mathbf w$} \HL{is zero-mean white Gaussian acceleration with a constant spectral-density matrix} \HL{$Q_d\succ0$}\HL{. Condition the prior on the endpoint position--velocity states} \HL{$\bsx^i=[\bsq^i;\dot{\bsq}^i]$} \HL{and} \HL{$\bsx^{i+1}=[\bsq^{i+1};\dot{\bsq}^{i+1}]$}\HL{. With} \HL{$s=(t-t_i)/h$}\HL{, its conditional position mean is}
\HL{\begin{equation}
\begin{aligned}
\bsq_r(t)&=(2s^3-3s^2+1)\bsq^i
 +(s^3-2s^2+s)h\dot{\bsq}^{i}\\
&\quad+(-2s^3+3s^2)\bsq^{i+1}
 +(s^3-s^2)h\dot{\bsq}^{i+1},
\end{aligned}
\label{eq:gp_hermite_execution}
\end{equation}}
\HL{and its conditional velocity mean is} \HL{$\dot{\bsq}_r(t)$}\HL{. Hence evaluating the prior's conditional mean at the inferred support states gives exactly the cubic Hermite reference used for execution.}
\end{proposition}

\begin{IEEEproof}
\HL{Set} \HL{$u=t-t_i$}\HL{. Let} \HL{$\boldsymbol\Phi(u)$} \HL{and} \HL{$Q(u)$} \HL{be the transition matrix and process covariance of Sec.~}\HL{\ref{sec:Dynamic_Factor}}\HL{, evaluated} over \HL{duration} \HL{$u$}\HL{, and define} \HL{$B(u)=Q(u)\boldsymbol\Phi(h-u)^\top Q(h)^{-1}$}\HL{. The Gaussian conditioning formula~}\HL{\cite{mukadamContinuoustimeGaussianProcess2018}} \HL{gives}
\HL{\begin{equation}
\bar{\bsx}(t)=\boldsymbol\Phi(u)\bsx^i+B(u)\big(\bsx^{i+1}-\boldsymbol\Phi(h)\bsx^i\big).
\label{eq:gp_bridge_mean}
\end{equation}}
\HL{In the position block, the constant} \HL{$Q_d$} \HL{cancels between the covariance product and inverse, yielding}
\HL{\begin{equation}\label{eq:gp_position_bridge_coefficients}
\begin{aligned}
[I,0]B(u)
&=\big[(3s^2-2s^3)I,\ h(s^3-s^2)I\big].
\end{aligned}
\end{equation}}
\HL{Substitution into the position block of Eq.~}\HL{\eqref{eq:gp_bridge_mean}} \HL{and collection of the four endpoint coefficients gives Eq.~}\HL{\eqref{eq:gp_hermite_execution}}\HL{. Its time derivative equals the velocity block of the conditional mean.}
\end{IEEEproof}

\HL{The interpolation map is linear in the endpoint states, so it also carries their inferred means} to \HL{the execution mean. For collision evaluation, the chain rule transfers the residual Jacobian at each interpolated state to both endpoints, allowing obstacle penalties inside an interval to adjust the endpoint positions and velocities. These denser evaluations improve the temporal resolution of collision assessment while retaining the same support-state dimension and local factor structure. With a common prior and consistent timestamps, collision factors, remote-state queries, and controller references therefore share the same optimized GP trajectory throughout planning, communication, and execution.}

\section{Simulations and Experiments} \label{sec:Simulations_and_Experiments}
\setlength{\textfloatsep}{4pt plus 1pt minus 1pt}
\setlength{\dbltextfloatsep}{4pt plus 1pt minus 1pt}
\setlength{\floatsep}{4pt plus 1pt minus 1pt}
\setlength{\dblfloatsep}{4pt plus 1pt minus 1pt}
\setlength{\intextsep}{4pt plus 1pt minus 1pt}
\setlength{\abovecaptionskip}{3pt}\setlength{\belowcaptionskip}{0pt}

\paragraph{Compared Method Design}
\HL{The baselines are Syn-GBP~\cite{jin2026distributedSafetyAware}, GPMP2~\cite{mukadamContinuoustimeGaussianProcess2018}, Gafur's NMPC~\cite{gafur2022dynamic}, DMPC~\cite{schwartingParallelAutonomyAutomated2017,ferrantiDistributedNonlinearTrajectory2023}, MR-STORM~\cite{evron2026mrstorm}, and NADMM~\cite{ferrantiDistributedNonlinearTrajectory2023}. Syn-GBP uses synchronous messages and adaptive support count at fixed $\Delta t=0.6$~s, without interpolation-time collision factors. Our multi-arm GPMP2 adaptation centrally solves a coupled graph using the official GTSAM optimization library; MR-STORM uses GPU-accelerated MPPI. We retain and tune core formulations for Panda/ Gazebo under shared scenes, targets, and execution limits (Supplement~S1).}

\paragraph{\HL{Experimental Settings}}
\HL{Figs.~\ref{fig:Framework} and~\ref{fig:collision_model}(c) show the simulation layouts. Method and mechanism comparisons use 50 matched long-horizon scenes: four 7-DoF Panda arms on a $2\times2$ grid with $0.8$~m nearest-neighbor spacing, facing the origin from identical fixed joint configurations. The minimum initial sphere-surface gap is $0.204$~m, leaving $0.104$~m beyond the $0.10$~m inter-arm safety-trigger threshold. Each arm is tasked with visiting three sequential joint-space targets within 60~s of simulation time while maintaining safety throughout execution. The $M=4$ spherical obstacles, with $r_{\mathrm{obs}}=0.08$~m and $v_{\mathrm{obs}}=0.20$~m/s, repeatedly rebound within the $2\times2$~m region $[-1,1]^2$.}

\HL{DD-GBP uses $N_{\mathrm{iter}}^{\max}=20$, $\epsilon_E=10^{-3}$, $\epsilon_c=0.10$~m, $r_c=0.4$~m, $N_{\mathrm{act}}^{\max}=8$ additional activations per trigger source, $N_{\min}=3$, and $\Delta t_k\in[0.01,0.6]$~s. Boundary-position and other factor weights are $\lambda_p^{\mathrm{init}}=10^4$, $\lambda_p^{\mathrm{goal}}=5000$, $\lambda_b=500$, $\lambda_o=\lambda_i=400$, and $\lambda_d$ from Eq.~\eqref{eq:adaptive_process_scaling}; the fixed-GP ablation uses $\lambda_d=125$. Prescribed velocity and acceleration bounds are $\rho=0.3$ times the Panda's physical limits. GP interpolation uses $n_{\mathrm{int}}=3$ internal monitoring points per interval, over $N_{\mathrm{obs}}^{\mathrm{int}}=3$ and $N_{\mathrm{inter}}^{\mathrm{int}}=2$ leading intervals, with $\lambda_o^{\mathrm{int}}=\lambda_o$ and $\lambda_i^{\mathrm{int}}=10\lambda_i$. These factors preserve the support-state dimension and sparse inference. Coarse-to-fine tuning selects fixed comparison parameters; settings and sensitivity plots are in Supplement~S1--S2. DD-GBP disables obstacle-state prediction by default.}

\HL{Simulation uses Gazebo/ ROS on an Intel Core i9-14900KF workstation. The GBP and optimization baselines use central processing unit (CPU) execution, whereas MR-STORM uses an NVIDIA GeForce RTX 4080 SUPER GPU with 16~GB memory. The physical platform consists of two Franka Emika Panda arms, a camera for obstacle detection at 30~Hz, and Transmission Control Protocol/ Internet Protocol (TCP/IP) communication with the robot controllers.}

\begin{figure*}[t]
\centering
\resizebox{\textwidth}{!}{%
\begin{tabular}{@{}c@{\hspace{5pt}}c@{\hspace{5pt}}c@{}}
\includegraphics[height=110pt]{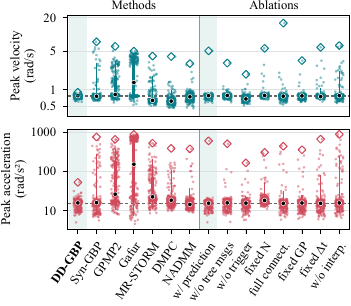} &
\includegraphics[height=110pt]{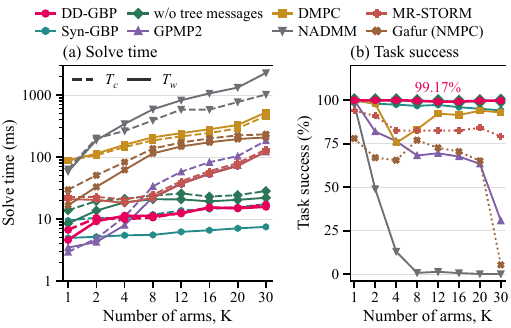} &
\includegraphics[height=110pt]{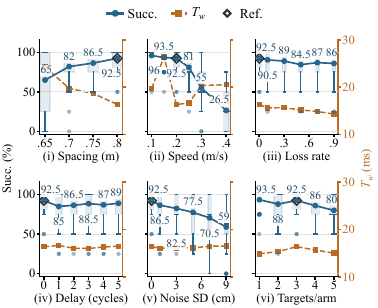} \\
\scriptsize (a) Executed kinematic peaks & \scriptsize (b) Obstacle-free scalability & \scriptsize (c) Scenario sensitivity
\end{tabular}
}
\caption{\HL{(a) Per-arm execution peaks in Tab.~\ref{tab:combined_results} order: circles show medians, diamonds show maxima. Gray lines mark DD-GBP's mean peaks. (b) Scaling: dashed/ solid curves show $T_{\mathrm c}$/ $T_{\mathrm w}$, alongside Succ. (c) Scenario sensitivity: boxes show per-scene Succ.; blue/ orange curves show mean Succ./ $T_{\mathrm w}$ (right axis); diamonds mark the reference.}}
\label{fig:scaling_time}
\label{fig:kinematic_peaks}
\makeatletter
\edef\@currentlabel{\thefigure(c)}\label{fig:scenario_ablation}
\makeatother
\end{figure*}

\begin{table*}[t]
\centering
\caption{\HL{50-scene methods (a) and ablations (b). w/ and w/o: with and without. Bold: group best.}}
\label{tab:combined_results}
\makeatletter
\edef\@currentlabel{\thetable(a)}\label{tab:batch_results}
\edef\@currentlabel{\thetable(b)}\label{tab:exp2_ablation_new}
\makeatother
\input{tables/std0927_comparison_paired.tex}
\end{table*}

\paragraph{Metrics}
\HL{The average per-arm performance metrics are:}
(1)~\HL{Task success \textbf{Succ.}: completion of all assigned targets within the run budget while maintaining $d(\bsq(t))>0$ throughout execution.}
(2)~\HL{Conditional clearance} $\bar d_{\mathrm s}=\frac{1}{|\Delta_T|}\int_{\Delta_T}d(\bsq(t))\,dt$, \HL{where $\Delta_T=\{t\in[0,T_{\mathrm{sum}}]:d(\bsq(t))\le\epsilon_c\}$ and $T_{\mathrm{sum}}$ is the elapsed execution time until all targets are reached or the run budget expires.}
(3)~\HL{Cold/ warm solve times $T_{\mathrm c}/T_{\mathrm w}$: mean solve times for the first planning call after target initialization or switching, and for calls with a warm initial trajectory, respectively.}
(4)~\HL{Goal count $C_{\mathrm g}$: the number of reached targets.}
(5)~\HL{Path length $L_{\mathrm g}$: joint-space trajectory length per reached goal.}
(6)~\HL{Time per goal $T_{\mathrm g}=T_{\mathrm{sum}}/C_{\mathrm g}$.}
(7)~\HL{Communication throughput $C_{\mathrm m}$: delivered trajectory-message bytes per arm per second.}
(8)~\HL{Tracking error $e_{\mathrm q}$: seven-joint position root-mean-square (RMS) error after reference-lag alignment; the underlying arm impedance controller introduces approximately $0.1$~s of filtering-related tracking lag.}
(9)~\HL{Peak velocity $v_{\mathrm{pk}}=\max_t\|\dot{\bsq}(t)\|_\infty$.}
(10)~\HL{Peak acceleration $a_{\mathrm{pk}}=\max_t\|\ddot{\bsq}(t)\|_\infty$, estimated by timestamp-aware velocity differences. Both peaks cover each arm's full execution window and are averaged across arms.}

\subsection{\HL{Exp.~I: Method Comparison and Scalability}}
\subsubsection{\HL{Method Comparison}}
\HL{DD-GBP leads the four-arm dynamic benchmark in Succ. ($92.5\%$), conditional clearance ($0.0735$~m), mean goal count ($3.00$), and measured tracking error ($2.41$~mrad; Tab.~\ref{tab:batch_results}). Its path length and time per goal are also near the best, without excessive detours to avoid neighboring arms and obstacles. Shorter paths and faster completion than the MPC/ MPPI baselines demonstrate efficient long-horizon progress while preserving safety. Supplementary videos demonstrate diverse multi-arm execution scenarios.}

\HL{Syn-GBP and GPMP2 omit interpolation-time collision factors, shortening solves but yielding only $53.5\%$ and $23.0\%$ success. Support-only checks can miss obstacles; GPMP2 can also settle in poor local solutions. Syn-GBP needs more cold iterations ($14.26\pm2.38$ versus DD-GBP's $10.61\pm3.69$). The schedule-only \emph{w/o tree messages} retains all other DD-GBP mechanisms yet is slower and less successful (Tab.~\ref{tab:exp2_ablation_new}, right).}

\HL{Gafur, DMPC, and NADMM require longer warm solves, follow longer paths, and spend more time per goal while reaching fewer goals. These results are consistent with slow updates hindering collision-constrained progress as obstacles move. MR-STORM reaches almost all targets with approximately $24$~ms native four-arm GPU updates, but its low Succ. is dominated by collisions. Its sampled policies provide insufficient collision-avoidance response in these rapidly changing scenes; fast native updates alone do not ensure safe execution.}

\HL{Fig.~\ref{fig:kinematic_peaks}(a) compares per-arm joint-velocity and acceleration peaks throughout execution. DD-GBP has the smallest upper-tail peaks and interquartile spreads among the baselines; its worst peaks of $0.900$~rad/s and $53.2$~rad/s$^2$ are markedly lower than Syn-GBP's $7.40$~rad/s and $761$~rad/s$^2$. Long planning times and slow response reduce DMPC's command-update frequency, helping explain its lower mean velocity peak; nevertheless, substantial extreme excursions remain. Adaptive planning and GP-consistent continuous execution suppress these transients in DD-GBP, enabling smoother, more stable execution.}

\subsubsection{\HL{Failure Analysis}}\label{sec:Failure_Analysis}
\HL{Of 200 DD-GBP arm runs, 14 contact moving obstacles; one further arm remains safe but is blocked from its final target by neighboring arms. No inter-arm collision occurs in these monitored runs. Isolated encounters generally leave room to maneuver; failures occur more often when several obstacles and arms interfere simultaneously, sharply restricting both avoidance space and response time. The success loss above $0.25$~m/s (Fig.~\ref{fig:scenario_ablation}) suggests that the $0.20$~m/s reference already imposes substantial dynamic difficulty under the prescribed motion limits. Even so, DD-GBP achieves $92.5\%$ collision-free all-target completion.}

\subsubsection{\HL{Scalability}}
\HL{Fig.~\ref{fig:scaling_time}(b) tests 1--30 arms without external obstacles in compact industrial-style workspaces ($1.0$~m nearest-neighbor spacing; 50 planned scenes per method/ size). DD-GBP retains $99.17$--$100\%$ success and $4.61$--$15.80$~ms warm solves. The w/o tree messages variant costs more at similar success. Interpolation-free Syn-GBP is faster for $K\ge2$, but large-team completion declines. GPMP2's centralized optimization slows sharply with state dimension and graph size. DMPC improves in larger teams but, like NADMM, remains computation-heavy. MR-STORM and Gafur show lower large-team completion.}

\subsection{\HL{Exp.~II: Mechanism and Scenario Ablations}}
\subsubsection{\HL{Mechanism Ablations}}
\HL{Tab.~\ref{tab:exp2_ablation_new} reports obstacle prediction and seven component ablations on the same 50 scenes, using DD-GBP in Tab.~\ref{tab:batch_results} as the shared reference. Removing support-count, time-step, or adaptive GP scaling fixes $N=10$, $\Delta t=0.6$~s, or $\lambda_d=125$; these fixed settings are also selected by tuning. The w/o subgraph variant sets $r_c=\infty$ and removes the neighbor cap for full-team connectivity, retaining tree-directed local sweeps. Other variants replace triggering with continual replanning or remove interpolation. Details are in Supplement~S1.}

\HL{Removing interpolation gives the lowest success ($64.0\%$) despite a $7.1$~ms warm solve. Fixing the time step or GP weight gives $77.5\%$ or $80.0\%$, demonstrating the value of temporal collision checks and adaptation. Without subgraph localization, redundant inter-arm links increase inference and communication work, with $83.5\%$ success and $4.82$~MB/s communication versus DD-GBP's $2.59$~MB/s. Continual replanning without triggering lowers success to $89.0\%$ and increases communication and time per goal. Together, interpolation, adaptation, localization, and selective replanning reinforce safety and task efficiency.}

\HL{The w/o tree messages variant records $90.0\%$ success, with $15.53\pm2.03$ cold and $10.19\pm1.66$ warm iterations, versus $10.61\pm3.69$ and $7.71\pm1.72$ for DD-GBP. With the same factor types and graph-construction rules, DD-GBP also records a lower normalized optimization loss ($\mathrm{Eng.}^{*}=1.071$ versus $4.216$), demonstrating the tree-directed schedule's solution quality and inference efficiency. The graph readily incorporates time-indexed obstacle-state prediction, achieving $98.0\%$ task success with $17.2$~ms warm solves and further improving safety and completion.}

\HL{As Fig.~\ref{fig:kinematic_peaks}(a) shows, removing adaptation, localized connectivity, or GP interpolation can reintroduce velocity or acceleration excursions. Nevertheless, these variants generally retain smaller typical peaks than Gafur's NMPC, illustrating the value of the common execution framework. The complete design combines tight peak distributions with higher task success; adding prediction preserves low typical peaks while further improving completion. These complementary mechanisms support smooth execution as well as effective coordination.}

\subsubsection{\HL{Scenario Sensitivity}}
\HL{Fig.~\ref{fig:scenario_ablation} reports task success and warm solve time in three groups of stress tests. \textbf{Extreme scenes:} spacing decreases to $0.65$~m, leaving only $0.054$~m initial inter-arm clearance, below the replanning margin; obstacle speed increases to $0.40$~m/s. DD-GBP still achieves $65.0\%$ success at the tightest spacing. The sharp decline above $0.25$~m/s indicates that these tightly shared scenes approach the tested configuration's dynamic-response limit under bounded arm speeds. Several fast obstacles and neighboring arms can jointly narrow the available space and leave insufficient time for avoidance. As discussed in Sec.~\ref{sec:Failure_Analysis}, higher arm-speed limits, wider spacing, or obstacle prediction could extend this response range.}

\HL{\textbf{Communication and sensing robustness:} success remains $86.0\%$ under $90\%$ packet loss and $85.0$--$89.0\%$ with one to five skipped send cycles, while warm solves remain $16.0$--$16.6$~ms. Even $5$~cm obstacle-position noise permits $77.5\%$ success. These results demonstrate robust coordination under stale neighbor information and substantial observation disturbances.}

\HL{\textbf{Long-horizon safety:} each arm continuously visits up to five sequential targets, with $80.0\%$ of arms completing every target without collision, compared with $93.5\%$ for one target. Sustained success through repeated goal switches demonstrates DD-GBP's ability to maintain safe, uninterrupted long-horizon execution.}

\subsection{Exp.~III: Real-World Multi-Robot Deployment}

\HL{We evaluate DD-GBP through transport and disturbance tests on two opposing Panda arms sharing a narrow worktable (supplementary videos). At $0.9$~m base spacing, their initial sphere-surface clearance is $0.180$~m, only $0.080$~m beyond the $0.10$~m planning margin. We model the tabletop as a planar ground obstacle and declare a target reached when $\|\bsq-\bsq_{\mathrm{goal}}\|_\infty\le0.04$~rad. The velocity coefficient is $0.2$; other planning parameters match the simulations, including the GP acceleration scale of $0.3$.}

\begin{figure}[!t]
 \centering
 \includegraphics[width=0.94\linewidth]{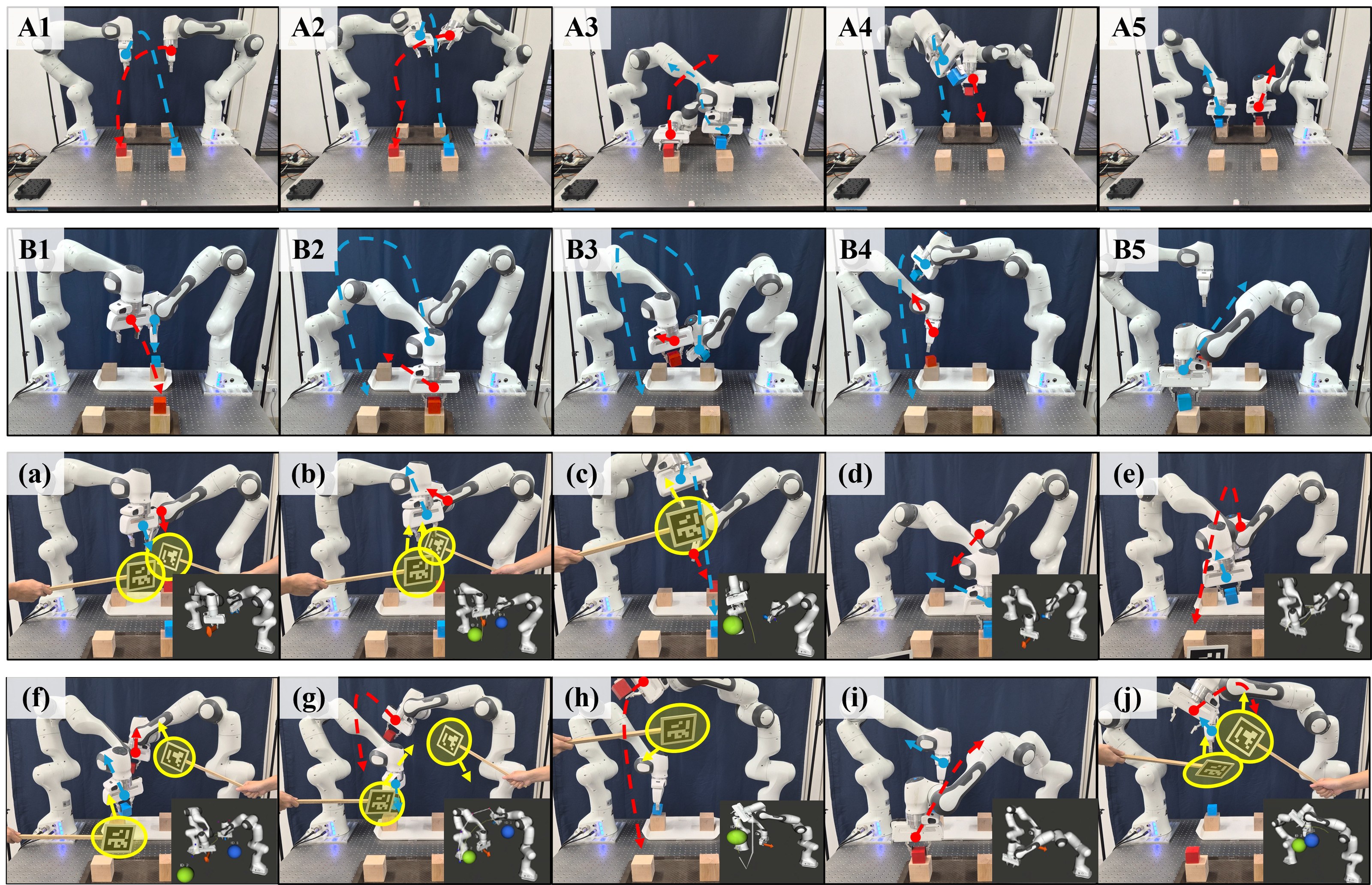}
 \caption{\HL{Intersecting transport (A1--A5, B1--B5) and multi-obstacle interference ((a)--(j)). Robot arrows indicate approximate future motion; yellow circles/ arrows mark detected obstacles/ disturbance directions; insets show the planning scene. Full executions appear in the supplementary videos.}}
 \label{fig:real_world_exp1}
\end{figure}

\subsubsection{Pick-and-Place Tasks in a Compact Workspace}

\HL{Given grasp, release, and return poses, DD-GBP plans paths and timing without prescribed avoidance or passage order. In Fig.~\ref{fig:real_world_exp1}, sequence A1--A5 shows front-to-rear transfers with closely spaced, collision-prone grasp configurations; B1--B5 shows staggered front/ rear crossing transfers requiring larger mutual-avoidance motions.}

\begin{figure}[!t]
 \centering
 \includegraphics[width=\linewidth]{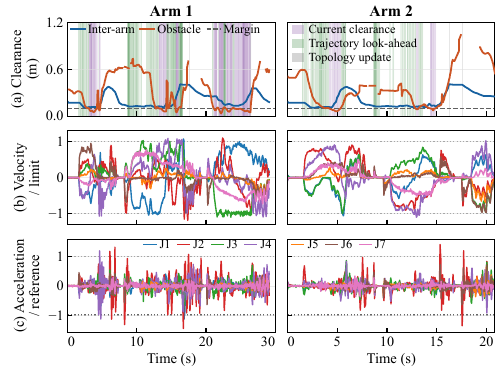}
 \caption{\HL{Execution under interference: (a) sphere-surface clearances and replanning triggers (shading); (b,c) joint velocities/ accelerations, normalized by per-joint velocity limits/ acceleration references (coefficients $0.2$/ $0.3$); dotted lines mark $\pm1$. Joint colors are shared by both arms.}}
 \label{fig:real_world_exp2}
 \label{fig:real_world_execution}
\end{figure}

\subsubsection{Dynamic Obstacle Avoidance in Shared Workspaces}

\HL{We introduce random manual disturbances during the B1--B5 task. A camera detects handheld ArUco markers at $30$~Hz; ROS publishes $0.08$~m-radius obstacle spheres (Fig.~\ref{fig:real_world_exp1}, (a)--(j)). Current-clearance and trajectory-look-ahead triggers update the remaining motion without obstacle-state prediction.}

\HL{In Fig.~\ref{fig:real_world_execution}(a), inter-arm clearance remains above the $0.10$~m planning margin, with a minimum of $0.107$~m, demonstrating safe separation during transport. Even under sustained hand-driven disturbances that follow the arms and temporarily reduce obstacle clearance below this margin, the arms adjust their motion and recover clearance, retaining positive minimum obstacle distances of $0.032$/ $0.060$~m.}

\HL{The trigger bands distinguish current-proximity responses from trajectory-look-ahead replanning. The planner evaluates the remaining GP trajectory against currently detected obstacles and revises affected future segments before the arm reaches them. This anticipatory trajectory query complements reactive avoidance without forecasting future obstacle states.}

\HL{Fig.~\ref{fig:real_world_execution}(b,c) shows per-joint velocities/ accelerations normalized by their corresponding velocity limits/ nominal acceleration references. Velocity changes during obstacle interference demonstrate responsive avoidance. Acceleration ratios stay near zero for most of the execution, indicating that accelerations generally remain below their references. Exceedance fractions are $1.44\%$ for velocity and $0.24\%$ for acceleration. These results demonstrate largely smooth, timely GP-consistent execution under sustained interference.}

\section{Conclusion}\label{sec:Conclusion}
DD-GBP \HL{combines switching factor-graph inference, adaptive horizons, and localized backward--forward messages} for online \HL{multi-arm} planning\HL{. A common GP representation connects support-state inference, interpolation-time factors, timestamp alignment, and continuous execution, with local replanning-consistency and early-stopping bounds under stated assumptions.}

\HL{On 50 four-arm scenes with four dynamic obstacles, DD-GBP records $92.5\%$ collision-free all-target completion and $16.3$~ms mean warm solves. Obstacle-free tests extend to 30 arms with $99.17$--$100\%$ single-arm task success and average planning times of order $10$~ms. Mechanism and scenario studies demonstrate robust long-horizon coordination under sensing and communication disturbances. Compact-workspace physical demonstrations further show the integrated architecture's deployability. Future work will advance nonlinear, time-varying GBP convergence theory and multi-robot coordination across heterogeneous platforms.}


\bibliographystyle{IEEEtran}
\bibliography{refs/refs.bib}


\clearpage
\setcounter{table}{0}\setcounter{figure}{0}\setcounter{equation}{0}
\renewcommand{\thetable}{S\arabic{table}}
\renewcommand{\thefigure}{S\arabic{figure}}
\renewcommand{\theequation}{S\arabic{equation}}
\renewcommand{\theHtable}{supp.\arabic{table}}
\renewcommand{\theHfigure}{supp.\arabic{figure}}
\renewcommand{\theHequation}{supp.\arabic{equation}}
\renewcommand{\theHsection}{supp.\arabic{section}}
\makeatletter\setlength{\@dblfptop}{0pt}\setlength{\@fptop}{0pt}\makeatother
\twocolumn[{\centering\Large\bfseries Supplementary Material\par
\vspace{4pt}\normalsize\normalfont
Benchmark Details, Additional Statistics, and Linearization Examples\par\vspace{12pt}}]
\input{supplementary.tex}
\end{document}

%% file: figures/ddgbp_framework_compact.tex
\def\ddgbpfigtext#1#2{\ifdefined\DDGBPChinese #2\else #1\fi}
\begin{tikzpicture}[
  x=1mm,y=.85mm,
  line cap=round,line join=round,
  block/.style={draw=black!70,line width=.4pt,rounded corners=1pt,
    text=black,align=center,text width=20.6mm,minimum height=6mm,
    inner xsep=1.7pt,inner ysep=1.5pt,
    font=\fontsize{7.2}{8.1}\selectfont,fill=white},
  flow/.style={draw=black!80,line width=.45pt,-{Latex[length=1.35mm,width=.85mm]}},
  exchange/.style={draw=blue,dashed,line width=.5pt,-{Latex[length=1.35mm,width=.85mm]}},
  event/.style={draw=black!65,dashed,line width=.45pt,-{Latex[length=1.35mm,width=.85mm]}},
  heading/.style={text=black,align=center,font=\fontsize{7.4}{8.3}\selectfont\bfseries},
  note/.style={text=black,align=center,font=\fontsize{6.8}{7.6}\selectfont}
]
\node[heading] at (30.75,73.5) {\ddgbpfigtext{DD-GBP planning}{DD-GBP 规划}};
\node[inner sep=0pt,anchor=north] at (77,75.5)
  {\includegraphics[width=18.3mm,trim=19.64bp 392.29bp 598.04bp 12.65bp,clip]{figures/optimized/robot_layout_1_q97.jpg}};
\node[note] at (77,52) {\ddgbpfigtext{Four-arm scene}{四臂场景}};
\node[heading,text width=19mm] at (77,46.5)
  {\ddgbpfigtext{Continuous\\execution}{连续执行}};
\node[block,fill=cyan!10,text width=53mm,minimum height=7mm]
  (rebuild) at (30.75,65.3)
  {\ddgbpfigtext{Dynamic subgraph reconstruction\\Adaptive time grid and initialization}{动态子图重构\\自适应时间网格与初始化}};

\begin{scope}[on background layer]
  \filldraw[fill=blue!4,draw=blue,line width=.5pt,rounded corners=1.5pt]
    (1.5,13.5) rectangle (60,59.3);
  \draw[draw=black!35,dashed,line width=.35pt,rounded corners=1pt]
    (3,17.4) rectangle (28.6,54);
  \draw[draw=black!35,dashed,line width=.35pt,rounded corners=1pt]
    (32,17.4) rectangle (59,54);
\end{scope}
\node[heading,text=blue] at (30.75,55.8)
  {\ddgbpfigtext{DD-GBP network inference}{DD-GBP 网络推断}};
\node[heading] at (15.8,52) {\ddgbpfigtext{Robot $k$}{机器人 $k$}};
\node[heading] at (44.8,52) {\ddgbpfigtext{Robot $j$}{机器人 $j$}};
\node[block,fill=blue!5] (krecv) at (15.8,46.5)
  {\ddgbpfigtext{Collect messages /\\observations}{收集消息／\\观测信息}};
\node[block,fill=cyan!10] (klin) at (15.8,38.2)
  {\ddgbpfigtext{Relinearize\\local factors}{重新线性化\\局部因子}};
\node[block,fill=orange!12] (kinf) at (15.8,29.9)
  {\ddgbpfigtext{Backward--forward\\Gaussian inference}{后向--前向\\高斯推断}};
\node[block,fill=green!10] (kpub) at (15.8,21.6)
  {\ddgbpfigtext{Publish belief /\\trajectory}{发布置信／\\轨迹消息}};
\node[block,fill=blue!5] (jrecv) at (44.8,46.5)
  {\ddgbpfigtext{Collect messages /\\observations}{收集消息／\\观测信息}};
\node[block,fill=cyan!10] (jlin) at (44.8,38.2)
  {\ddgbpfigtext{Relinearize\\local factors}{重新线性化\\局部因子}};
\node[block,fill=orange!12] (jinf) at (44.8,29.9)
  {\ddgbpfigtext{Backward--forward\\Gaussian inference}{后向--前向\\高斯推断}};
\node[block,fill=green!10] (jpub) at (44.8,21.6)
  {\ddgbpfigtext{Publish belief /\\trajectory}{发布置信／\\轨迹消息}};
\draw[flow] (rebuild.south) -- (30.75,59.3);
\foreach \r in {k,j}{
  \draw[flow] (\r recv) -- (\r lin);
  \draw[flow] (\r lin) -- (\r inf);
  \draw[flow] (\r inf) -- (\r pub);
}
\draw[flow] (kpub.west) -- (3.8,21.6) |- (krecv.west);
\draw[flow] (jpub.east) -- (58.1,21.6) |- (jrecv.east);
\draw[exchange] (kpub.east) -- (jrecv.west);
\fill[blue!4] (30.3,34.05) circle[radius=.4mm];
\draw[exchange] (jpub.west) -- (krecv.east);
\node[note] at (30.75,15.3)
  {\ddgbpfigtext{Iterate: local stop / budget}{迭代：局部停止准则／预算}};

\node[block,text width=17mm,fill=green!10] (reference) at (77,36.9)
  {\ddgbpfigtext{GP / Hermite\\reference}{GP／Hermite\\连续参考}};
\node[block,text width=17mm,fill=orange!12] (tracking) at (77,26.7)
  {\ddgbpfigtext{Low-level\\tracking}{底层跟踪}};
\node[block,text width=17mm,fill=black!5] (robots) at (77,16.5)
  {\ddgbpfigtext{Robots}{机器人}};
\draw[flow] (60,19.5) -- (61.8,19.5) |- (reference.west);
\node[note,rotate=90] at (64.5,28)
  {\ddgbpfigtext{Current plan}{当前轨迹}};
\draw[flow] (reference) -- (tracking);
\draw[flow] (tracking) -- (robots);

\node[block,text width=75mm,minimum height=7.5mm,fill=orange!12]
  (triggers) at (45,5.2)
  {\ddgbpfigtext{Replanning triggers\\Predicted collisions, new goals, neighborhood changes}{重规划触发\\碰撞预测、新目标、邻域变化}};
\coordinate (plan-monitor) at (15.5,13.5);
\draw[event] (plan-monitor) -- (plan-monitor |- triggers.north);
\draw[event] (robots.south) -- (robots.south |- triggers.north);
\draw[event] (triggers.west) -- (0,5.2) |- (rebuild.west);
\end{tikzpicture}

%% file: tables/std0927_comparison_paired.tex
\begin{minipage}[t]{0.465\textwidth}
\centering\scriptsize
\setlength{\tabcolsep}{1.10pt}\renewcommand{\arraystretch}{1.12}
\begin{NoHyper}
\begin{tabular}{@{}lrrrrrrrrrr@{}}
\toprule
(a) Method & Succ. $\uparrow$ & $\bar d_{\mathrm s}\uparrow$ & $T_{\mathrm c}\downarrow$ & $T_{\mathrm w}\downarrow$ & $C_{\mathrm g}\uparrow$ & $L_{\mathrm g}\downarrow$ & $T_{\mathrm g}\downarrow$ & \HL{$e_{\mathrm q}\downarrow$} & \HL{$v_{\mathrm{pk}}\downarrow$} & \HL{$a_{\mathrm{pk}}\downarrow$} \\
 & (\%) & (mm) & (ms) & (ms) &  & (rad) & (s) & \HL{(mrad)} & \HL{(rad/s)} & \HL{(rad/s$^2$)} \\
\midrule
\textbf{DD-GBP} & \HL{\textbf{92.5}} & \HL{\textbf{73.5}} & \HL{25.8} & \HL{16.3} & \HL{\textbf{3.00}} & \HL{4.14} & \HL{5.88} & \HL{\textbf{2.41}} & \HL{0.77} & \HL{\textbf{15.6}} \\
Syn-GBP & \HL{53.5} & \HL{58.0} & \HL{\textbf{15.5}} & \HL{\textbf{7.5}} & \HL{2.99} & \HL{\textbf{4.04}} & \HL{\textbf{5.77}} & \HL{5.26} & \HL{1.0} & \HL{48.3} \\
\HL{GPMP2} & \HL{23.0} & \HL{29.9} & \HL{22.6} & \HL{15.1} & \HL{2.93} & \HL{4.10} & \HL{7.13} & \HL{15.81} & \HL{1.4} & \HL{109.8} \\
\HL{Gafur} & \HL{19.5} & \HL{48.9} & \HL{111.0} & \HL{88.0} & \HL{2.05} & \HL{11.78} & \HL{27.51} & \HL{7.59} & \HL{2.2} & \HL{357.5} \\
\HL{MR-STORM} & \HL{9.5} & \HL{25.4} & \HL{24.3} & \HL{24.1} & \HL{2.95} & \HL{5.32} & \HL{13.14} & \HL{6.86} & \HL{0.81} & \HL{59.4} \\
DMPC & \HL{2.0} & \HL{39.9} & \HL{191.8} & \HL{229.0} & \HL{2.24} & \HL{6.47} & \HL{28.42} & \HL{4.53} & \HL{\textbf{0.67}} & \HL{28.8} \\
NADMM & \HL{0.0} & \HL{37.0} & \HL{607.2} & \HL{559.2} & \HL{1.05} & \HL{11.46} & \HL{56.75} & \HL{4.41} & \HL{0.78} & \HL{22.8} \\
\multicolumn{11}{@{}p{\linewidth}@{}}{\scriptsize\HL{DD-GBP data in (a) also serve as (b)'s reference; $C_{\mathrm m}=2.59$ MB/s.}} \\
\bottomrule
\end{tabular}
\end{NoHyper}
\end{minipage}
\hfill
\begin{minipage}[t]{0.520\textwidth}
\centering\scriptsize
\setlength{\tabcolsep}{1.10pt}\renewcommand{\arraystretch}{1.12}
\begin{NoHyper}
\begin{tabular}{@{}lrrrrrrrrrrr@{}}
\toprule
(b) Variant & Succ. $\uparrow$ & $\bar d_{\mathrm s}\uparrow$ & $T_{\mathrm c}\downarrow$ & $T_{\mathrm w}\downarrow$ & $C_{\mathrm g}\uparrow$ & $L_{\mathrm g}\downarrow$ & $T_{\mathrm g}\downarrow$ & \HL{$C_{\mathrm m}\downarrow$} & \HL{$e_{\mathrm q}\downarrow$} & \HL{$v_{\mathrm{pk}}\downarrow$} & \HL{$a_{\mathrm{pk}}\downarrow$} \\
 & (\%) & (mm) & (ms) & (ms) &  & (rad) & (s) & \HL{(MB/s)} & \HL{(mrad)} & \HL{(rad/s)} & \HL{(rad/s$^2$)} \\
\midrule
\HL{\textbf{w/ obs pred.}} & \HL{\textbf{98.0}} & \HL{\textbf{87.4}} & \HL{25.8} & \HL{17.2} & \HL{2.99} & \HL{4.10} & \HL{5.88} & \HL{2.64} & \HL{2.43} & \HL{0.79} & \HL{17.8} \\
\HL{w/o tree msgs.} & \HL{90.0} & \HL{72.4} & \HL{34.8} & \HL{20.7} & \HL{\textbf{3.00}} & \HL{4.14} & \HL{5.85} & \HL{2.73} & \HL{2.53} & \HL{0.80} & \HL{19.0} \\
\HL{w/o trigger} & \HL{89.0} & \HL{72.7} & \HL{25.9} & \HL{13.3} & \HL{\textbf{3.00}} & \HL{4.17} & \HL{6.26} & \HL{2.78} & \HL{\textbf{1.98}} & \HL{\textbf{0.70}} & \HL{\textbf{16.9}} \\
\HL{w/o adapt. $N$} & \HL{87.0} & \HL{72.1} & \HL{24.1} & \HL{18.9} & \HL{\textbf{3.00}} & \HL{4.37} & \HL{7.43} & \HL{3.64} & \HL{4.60} & \HL{0.83} & \HL{25.2} \\
\HL{w/o subgraph} & \HL{83.5} & \HL{70.9} & \HL{37.9} & \HL{23.9} & \HL{\textbf{3.00}} & \HL{4.30} & \HL{6.31} & \HL{4.82} & \HL{2.76} & \HL{0.84} & \HL{22.2} \\
\HL{w/o GP scaling} & \HL{80.0} & \HL{69.0} & \HL{18.3} & \HL{12.9} & \HL{\textbf{3.00}} & \HL{\textbf{3.97}} & \HL{6.00} & \HL{2.49} & \HL{2.11} & \HL{0.77} & \HL{20.6} \\
\HL{w/o adapt. $\Delta t$} & \HL{77.5} & \HL{67.4} & \HL{20.2} & \HL{14.1} & \HL{\textbf{3.00}} & \HL{4.37} & \HL{\textbf{5.73}} & \HL{\textbf{2.40}} & \HL{2.30} & \HL{0.83} & \HL{26.9} \\
\HL{w/o interp.} & \HL{64.0} & \HL{62.2} & \HL{\textbf{12.8}} & \HL{\textbf{7.1}} & \HL{2.99} & \HL{4.02} & \HL{5.96} & \HL{4.00} & \HL{4.35} & \HL{0.96} & \HL{49.8} \\
\bottomrule
\end{tabular}
\end{NoHyper}
\end{minipage}

%% file: supplementary.tex
\begingroup\color{black}
\section*{S1. Benchmark protocol and implementations}\label{sec:supp_resource}\label{sec:supp_benchmark}
The simulation benchmarks and repeated physical experiments share the DD-GBP planning mechanisms below. Physical joint velocities use coefficient $0.2$, while the GP acceleration scale remains $0.3$. Manuscript Exp.~III and Figs.~\ref*{fig:real_world_exp1}--\ref*{fig:real_world_execution} describe the opposing arms at $0.9$~m base spacing, $30$~Hz ArUco detection, $0.08$~m obstacle spheres, and execution without obstacle prediction.

\subsubsection*{Scenes and fair comparison}
The method comparison and mechanism ablations use the same 50 seeded scenes: four seven-joint Panda arms at $(\pm0.4,\pm0.4,0)$~m, facing the origin, with $0.8$~m nearest-neighbor spacing. Initial joint configurations are identical and fixed. Each arm visits three randomized joint-space targets, screened at $0.2$~m clearance; completed arms remain as obstacles. Four spherical obstacles of radius $0.08$~m move at height $0.8$~m in $[-1,1]^2$~m with speed $0.20$~m/s, continuously rebounding at the region boundaries.

We retain each baseline's core formulation and use the authors' released implementations where available, adapting robot geometry and simulation interfaces to Panda/ ROS/ Gazebo. All methods receive careful hyperparameter tuning under common scenes, targets, nominal execution limits, and external performance metrics. Their native planning and execution mechanisms are retained rather than replaced by DD-GBP components.

Scenario sweeps change only the listed base spacing, obstacle speed, packet-loss rate, belief-update delay, observation noise, or target count. At $0.65$~m spacing, the initial-clearance screening margin is relaxed to generate starting scenes, without changing the signed-clearance evaluation. Obstacle-free scalability uses three targets per arm and $K\in\{1,2,4,8,12,16,20,30\}$, with 50 scenes planned per method/ size. Multi-arm rectangular grids have $1.0$~m nearest-neighbor spacing and face the origin; the single arm is placed at the origin.

\subsubsection*{Computing and execution}
Experiments run on an Intel Core i9-14900KF workstation; MR-STORM uses an NVIDIA GeForce RTX 4080 SUPER GPU with 16~GB memory, and the other solvers use the CPU. The simulation benchmarks use nominal velocity and acceleration coefficients of $0.3$ relative to Panda's physical limits; the physical demonstrations use velocity coefficient $0.2$. Each mission has a 60~s simulation-time budget; an arm that does not complete all targets within this budget is counted as a task failure.

\begin{table}[htbp]
\centering\fontsize{9}{10.8}\selectfont\setlength{\tabcolsep}{3pt}\color{black}
\caption{DD-GBP reference settings.}\label{tab:supp_frozen}
\begin{tabular}{@{}>{\raggedright\arraybackslash}p{0.30\linewidth}>{\raggedright\arraybackslash}p{0.64\linewidth}@{}}
\toprule
Component & Setting\\
\midrule
Time grid & Joint support-count/ interval adaptation; $N_{\min}=3$, $N_{\max}=14$, $\Delta t\in[0.01,0.6]$~s\\
Inference & Backward--forward scheduling; 20-iteration cap; objective-change tolerance $10^{-3}$\\
Replanning & Event-triggered; GP-resampled warm starts; linear initialization for a new target\\
Collision model & Spheres; margin $0.10$~m; obstacle aggregation $\alpha=30$\\
Collision precision & Obstacle/ inter-arm support-state precision 400; inter-arm interpolation multiplier 10\\
Interpolation & Three internal points per interval; first three obstacle intervals and first two inter-arm intervals\\
Boundary position precision & Start $10^4$; target 5000\\
Goal arrival & Maximum absolute joint-position error $\le0.07$~rad\\
Adaptive GP scaling & $Q_d=\lambda_d^{-1}\mathbf I$, $\lambda_d(\Delta t)=\sigma_z^2/(\bar a^2\Delta t)$, $\sigma_z=3$; fixed-scale ablation: $\lambda_d=125$\\
Interaction & $r_c=0.4$~m sphere-clearance gate; at most eight neighbors and eight additional activations per trigger source\\
Prediction & Disabled by default; measured-velocity forecasting only in its labeled variant\\
\bottomrule
\end{tabular}
\end{table}

\subsubsection*{Comparison methods}
\textbf{Syn-GBP} follows our earlier planner~\cite{jin2026distributedSafetyAware}. It uses synchronous messages without interpolation-time collision factors, and adapts the support count at fixed $\Delta t=0.6$~s. The \emph{w/o tree messages} ablation instead retains all current DD-GBP mechanisms and changes only the message schedule.

\textbf{GPMP2} is our centralized multi-arm adaptation of the GPMP2 formulation~\cite{mukadamContinuoustimeGaussianProcess2018}. It retains the continuous-time GP prior and maximum a posteriori factor-graph objective, coupling all arms through inter-arm collision factors. We implement the platform-specific factors and use the official \href{https://github.com/borglab/gtsam}{GTSAM library}~\cite{dellaert2012gtsam} with its native Levenberg--Marquardt optimizer, rather than GBP message inference. Panda geometry, inter-arm geometry queries, and ROS/ Gazebo interfaces provide the multi-arm adaptation; the core GP model and numerical optimizer are retained and the parameters are tuned. The tested configuration uses full inter-arm connectivity and adaptive support count at fixed $\Delta t=0.6$~s, without interpolation-time collision factors. The authors' \href{https://github.com/gtrll/gpmp2}{GPMP2 project} and \href{https://github.com/borglab/gpmp2}{maintained repository} provide the reference formulation.

\textbf{DMPC} follows the core distributed predictive-optimization formulations in~\cite{schwartingParallelAutonomyAutomated2017,ferrantiDistributedNonlinearTrajectory2023}: joint-velocity inputs, single-integrator dynamics, and linearized sphere-clearance constraints with nonnegative slack, solved by CasADi/ IPOPT. Careful tuning selects $N=15$, $\Delta t=0.08$~s, five outer updates, and collision-slack/ state-increment/ goal weights 50/20/50.

\textbf{NADMM} follows the local predictive subproblems and inter-arm consensus formulation in~\cite{ferrantiDistributedNonlinearTrajectory2023}. Its tuned settings are $N=10$, $\Delta t=0.08$~s, five ADMM updates, penalty $\rho=0.01$, relaxation $\beta=0.5$, and collision/ state-increment/ goal weights 50/20/50.

\textbf{MR-STORM}~\cite{evron2026mrstorm} uses the authors' \href{https://github.com/RoboWorkshop/rl_for_curobo}{official open-source cuRobo/ MPPI implementation}, retaining acceleration-policy sampling, peer-intention exchange, distance-dependent priorities, and the released diagonal distribution update. Tuned settings are 400 samples, two MPPI updates, ten steps at $\Delta t=0.1$~s, and a $0.1$~s replanning period. Policies are computed in a shared GPU batch with CUDA Graph capture disabled; the selected next position is sent to the tracking interface. Panda geometry, joint goals, collision costs, and ROS transport provide the benchmark integration; external obstacle positions are held over each horizon.

\textbf{Gafur (NMPC)}~\cite{gafur2022dynamic} builds on the \href{https://github.com/ngafur/MPC_MultiRobot_Motion_Planning}{authors' open-source Python implementation}, retaining joint-acceleration inputs, double-integrator prediction, ellipsoid--line-segment collision constraints, and deadlock coordination. Tuned settings are six steps at $\Delta t=0.2$~s, a $0.2$~s replanning period, ten IPOPT iterations, JIT, MUMPS, projection slope 10, and knot-level collision constraints. Planning acceleration bounds use nominal Panda limits and velocity bounds use coefficient $0.3$. A $0.8$~s execution prefix uses \texttt{FollowJointTrajectory} with velocity-limited integration. Neighbor policies are cached and shifted by the author-derived cycle rule, with a two-phase initial exchange. Panda geometry, seven-joint weights, and ROS interfaces supply the platform adaptation.

\subsubsection*{Subgraph and delay ablations}
The \emph{w/o subgraph} variant sets $r_c=\infty$ and removes the separate neighbor cap, giving full-team connectivity. Tree-directed sweeps, triggering, interpolation, the adaptive grid, and the eight-additional-arm activation budget remain unchanged. The extraction routine still runs but no longer localizes the four-arm team.

For delay $N=1,\ldots,5$, each established directed link skips $N$ send opportunities, holds its previous belief, and then receives a fresh belief. This models stale updates rather than a fixed millisecond network latency; graph reconstruction and shared-state neighbor discovery retain their normal behavior.

\subsubsection*{Cold/ warm calls}
$T_{\mathrm c}$ averages the first solves after target initialization or switching, and $T_{\mathrm w}$ averages subsequent calls, first within arms and then across arms. MR-STORM retains GPU-batch durations; Gafur distinguishes non-shifted and shifted initialization through its native interface.

\section*{S2. Additional results and parameter sensitivity}\label{sec:supp_statistics}
We provide parameter sensitivity, detailed scalability/ scenario statistics, and compact tracking/ execution diagnostics.

\subsection*{Parameter sensitivity and selection}\label{sec:supp_hyperparameters}
\begingroup\small
Nineteen settings vary one parameter at a time around the fixed DD-GBP configuration on the same 50 four-arm, three-target scenes. The four axes are interaction clearance $r_c$, obstacle-distance aggregation $\alpha$, shared obstacle/ inter-arm precision $\lambda_{\mathrm c}$, and the leading obstacle-interpolation budget $m_{\mathrm o}$. Defaults are $0.4$~m, 30, 400, and three intervals, respectively. Fig.~\ref{fig:supp_hyperparameters} and Tab.~\ref{tab:supp_hyperparameters} share the current manuscript DD-GBP reference: $92.5\%$ Succ., $25.753$~ms cold and $16.346$~ms warm solve time.

\begin{figure}[!t]
\centering
\includegraphics[width=\linewidth]{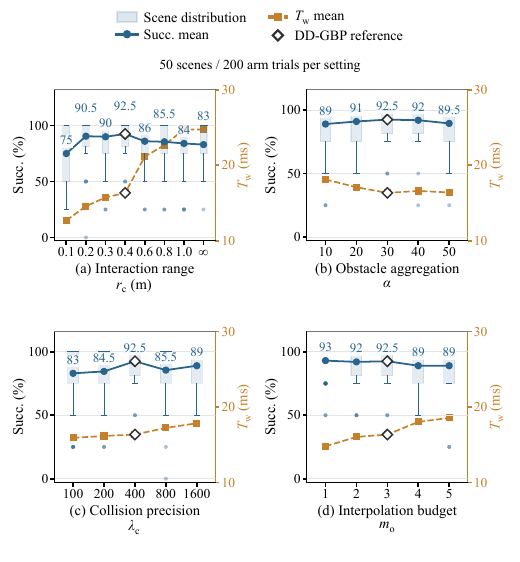}
\caption{Single-parameter sensitivity: boxes summarize per-scene Succ.; blue/ orange curves show mean Succ./ warm solve time; diamonds mark the shared DD-GBP reference. $\lambda_{\mathrm c}$ changes obstacle and inter-arm precisions together; $\infty$ removes the interaction-range restriction.}\label{fig:supp_hyperparameters}
\end{figure}

Coarse-to-fine selection first identifies useful operating ranges, then refines the safety/ computation trade-off and freezes the configuration. The sweeps show a broad favorable region: $\alpha=10$--$50$ retains $89.0$--$92.5\%$ success with $16.3$--$18.1$~ms warm solves; $r_c=0.2$--$0.4$~m retains $90.0$--$92.5\%$ success. Smaller neighborhoods reduce anticipatory coordination, while larger ones increase computation without improving success. Collision precision is more sensitive. One to five interpolation intervals retain $89.0$--$93.0\%$ success with $14.8$--$18.6$~ms warm solves. These results favor balanced tuning rather than uniformly increasing weights or budgets.

\begin{table}[!t]
\centering\fontsize{8.5}{10}\selectfont\setlength{\tabcolsep}{2pt}\color{black}
\caption{Current one-parameter studies: 50 scenes and 200 arms per setting; one shared reference.}\label{tab:supp_hyperparameters}
\begin{tabular*}{\linewidth}{@{\extracolsep{\fill}}llrrr@{}}
\toprule
Parameter & Value & \shortstack{Succ. $\uparrow$\\(\%)} & \shortstack{$T_{\mathrm c}$ $\downarrow$\\(ms)} & \shortstack{$T_{\mathrm w}$ $\downarrow$\\(ms)}\\
\midrule
DD-GBP (ref.) & Default & 92.5 & 25.753 & 16.346 \\
Range $r_{\mathrm c}$ (m) & 0.1 & 75.0 & 11.962 & 12.698 \\
Range $r_{\mathrm c}$ (m) & 0.2 & 90.5 & 15.078 & 14.572 \\
Range $r_{\mathrm c}$ (m) & 0.3 & 90.0 & 24.065 & 15.750 \\
Range $r_{\mathrm c}$ (m) & 0.6 & 86.0 & 33.444 & 21.128 \\
Range $r_{\mathrm c}$ (m) & 0.8 & 85.5 & 35.430 & 22.675 \\
Range $r_{\mathrm c}$ (m) & 1 & 84.0 & 38.453 & 24.729 \\
Range $r_{\mathrm c}$ (m) & $\infty$ & 83.0 & 37.489 & 24.728 \\
Aggregation $\alpha$ & 10 & 89.0 & 25.631 & 18.123 \\
Aggregation $\alpha$ & 20 & 91.0 & 25.768 & 17.094 \\
Aggregation $\alpha$ & 40 & 92.0 & 24.820 & 16.598 \\
Aggregation $\alpha$ & 50 & 89.5 & 25.239 & 16.405 \\
Precision $\lambda_{\mathrm c}$ & 100 & 83.0 & 24.966 & 15.909 \\
Precision $\lambda_{\mathrm c}$ & 200 & 84.5 & 25.007 & 16.159 \\
Precision $\lambda_{\mathrm c}$ & 800 & 85.5 & 25.886 & 17.231 \\
Precision $\lambda_{\mathrm c}$ & 1600 & 89.0 & 25.903 & 17.838 \\
Interp. budget $m_{\mathrm o}$ & 1 & 93.0 & 23.373 & 14.797 \\
Interp. budget $m_{\mathrm o}$ & 2 & 92.0 & 23.963 & 16.029 \\
Interp. budget $m_{\mathrm o}$ & 4 & 89.0 & 27.759 & 18.018 \\
Interp. budget $m_{\mathrm o}$ & 5 & 89.0 & 28.268 & 18.551 \\
\bottomrule
\end{tabular*}
\end{table}
\endgroup

\subsection*{Scalability and scenario results}
Tabs.~\ref{tab:supp_scaling_compact}--\ref{tab:supp_scenarios_compact} detail the scalability and scenario studies in manuscript Fig.~\ref*{fig:scaling_time}(b,c). Seven task/ planning metrics are retained; tracking error and kinematic peaks are reported separately. DD-GBP maintains high completion and short warm solves as the team grows, while the scenario sweeps identify its robustness and dynamic-response limits.

\begin{table*}[!p]
\centering
\noindent\begin{minipage}[t]{0.485\textwidth}
\input{tables/supp_scaling_results.tex}
\end{minipage}\hfill
\begin{minipage}[t]{0.485\textwidth}
\input{tables/supp_scenario_results.tex}

\par\vspace{6pt}
\begingroup\small\setlength{\parskip}{2pt}
\subsection*{Tracking and execution quality}\label{sec:supp_execution_tracking}
The compact comparison retains aligned and unshifted joint-position root-mean-square (RMS) errors, velocity RMS/ peak, and acceleration RMS/ peak. Lag alignment uses a configuration-wide lag fitted on separate scene folds; filtering in the impedance controller introduces an approximately $0.1$~s response lag. Gafur uses its joint-trajectory controller's desired reference.

\par
DD-GBP combines low aligned tracking error with the smallest extreme execution peaks among the compared methods: $0.900$~rad/s and $53.2$~rad/s$^2$, versus Syn-GBP's $7.40$~rad/s and $760.7$~rad/s$^2$. Its adaptive planning and GP-consistent continuous execution suppress abrupt velocity and acceleration transients, producing tighter peak distributions while maintaining high task completion (manuscript Tab.~\ref*{tab:batch_results} and Fig.~\ref*{fig:kinematic_peaks}(a)). DMPC's slow planning and response reduce its update frequency and typical motion speed, but large extreme peaks remain. Removing adaptation, localized connectivity, or interpolation can reintroduce execution excursions; the full framework improves execution stability rather than merely lowering average motion speed.

\endgroup

\end{minipage}

\end{table*}

\endgroup

\begin{table*}[!p]
\input{tables/supp_execution_diagnostics.tex}
\vspace{8pt}
\begin{minipage}{\textwidth}
\begingroup\normalsize
\input{collision_math.tex}

\endgroup
\end{minipage}
\end{table*}

%% file: tables/supp_scaling_results.tex
\providecommand{\CompactMethodHeader}{Method}
\providecommand{\CompactSampleHeader}{$K$}
\providecommand{\CompactScalingCaption}{Detailed scalability results for Fig.~\ref*{fig:scaling_time}(b).}
\providecommand{\CompactScalingNote}{Runs ending before task initialization are excluded. If a group has no near-contact interval, $\bar d_s=100$~mm denotes the evaluation cap. GPMP2 at $K=1$: warm times use only arms with warm calls.}
\begingroup
\fontsize{8}{8.6}\selectfont
\setlength{\tabcolsep}{1.1pt}
\renewcommand{\arraystretch}{0.90}

\captionof{table}{\CompactScalingCaption}\label{tab:supp_scaling_compact}
\noindent\begin{tabular*}{\linewidth}{@{\extracolsep{\fill}}lrrrrrrr@{}}
\toprule
\CompactSampleHeader & Succ.$\uparrow$ & $\bar d_s\uparrow$ & $T_c\downarrow$ & $T_w\downarrow$ & $C_g\uparrow$ & $L_g\downarrow$ & $T_g\downarrow$ \\
 & (\%) & (mm) & (ms) & (ms) &  & (rad) & (s) \\
\midrule
\multicolumn{8}{l}{\textbf{DD-GBP}} \\
1 & 100.0 & 100.0 & 6.7 & 4.6 & 3.00 & 3.79 & 4.39 \\
2 & 100.0 & 100.0 & 10.4 & 9.2 & 3.00 & 3.80 & 4.54 \\
4 & 100.0 & 92.7 & 10.0 & 11.3 & 3.00 & 3.77 & 4.73 \\
8 & 99.8 & 94.8 & 10.6 & 11.2 & 3.00 & 3.76 & 4.83 \\
12 & 99.2 & 93.3 & 12.6 & 12.3 & 2.99 & 3.79 & 4.85 \\
16 & 99.3 & 94.1 & 15.2 & 15.4 & 2.99 & 3.80 & 4.96 \\
20 & 99.7 & 94.3 & 15.1 & 14.9 & 3.00 & 3.80 & 4.90 \\
30 & 99.9 & 93.3 & 17.1 & 15.8 & 3.00 & 3.79 & 4.84 \\
\addlinespace[1pt]
\multicolumn{8}{l}{\textbf{Syn-GBP}} \\
1 & 100.0 & 87.4 & 9.5 & 5.0 & 3.00 & 3.84 & 4.77 \\
2 & 98.0 & 82.4 & 10.5 & 5.2 & 2.98 & 3.88 & 5.37 \\
4 & 97.5 & 80.9 & 10.3 & 5.5 & 2.96 & 3.89 & 6.15 \\
8 & 96.8 & 84.2 & 11.8 & 5.5 & 2.96 & 3.83 & 5.53 \\
12 & 97.3 & 83.6 & 13.7 & 6.2 & 2.98 & 3.86 & 5.49 \\
16 & 96.0 & 81.2 & 14.5 & 6.6 & 2.95 & 3.86 & 5.71 \\
20 & 95.1 & 82.4 & 15.6 & 7.1 & 2.96 & 3.86 & 5.52 \\
30 & 94.2 & 79.9 & 17.4 & 7.5 & 2.94 & 3.89 & 5.97 \\
\addlinespace[1pt]
\multicolumn{8}{l}{\textbf{GPMP2}} \\
1 & 100.0 & 66.7 & 2.9 & 3.5 & 3.00 & 3.88 & 4.18 \\
2 & 82.0 & 46.2 & 4.9 & 4.3 & 2.97 & 3.98 & 5.43 \\
4 & 76.5 & 51.1 & 10.9 & 8.0 & 2.97 & 3.99 & 6.39 \\
8 & 68.3 & 47.7 & 33.8 & 22.4 & 2.96 & 3.98 & 7.53 \\
12 & 69.5 & 50.7 & 57.9 & 37.8 & 2.91 & 4.05 & 9.15 \\
16 & 67.6 & 50.7 & 82.6 & 54.5 & 2.92 & 4.01 & 9.00 \\
20 & 63.4 & 49.5 & 103.7 & 74.1 & 2.86 & 4.12 & 11.43 \\
30 & 30.7 & 47.9 & 184.0 & 121.7 & 2.30 & 5.00 & 21.56 \\
\addlinespace[1pt]
\multicolumn{8}{l}{\textbf{w/o tree messages}} \\
1 & 100.0 & 100.0 & 13.9 & 8.5 & 3.00 & 3.79 & 4.39 \\
2 & 100.0 & 99.1 & 19.5 & 13.7 & 3.00 & 3.79 & 4.55 \\
4 & 100.0 & 88.5 & 21.2 & 18.6 & 3.00 & 3.76 & 4.75 \\
8 & 99.5 & 88.4 & 23.8 & 20.9 & 3.00 & 3.76 & 4.85 \\
12 & 99.5 & 88.3 & 25.9 & 20.8 & 2.99 & 3.79 & 4.83 \\
16 & 99.1 & 93.8 & 23.0 & 19.3 & 2.99 & 3.79 & 4.99 \\
20 & 99.5 & 88.9 & 24.3 & 20.4 & 3.00 & 3.79 & 4.92 \\
30 & 99.8 & 89.2 & 28.3 & 22.1 & 3.00 & 3.78 & 4.86 \\
\addlinespace[1pt]
\multicolumn{8}{l}{\textbf{DMPC}} \\
1 & 100.0 & 99.9 & 88.3 & 89.0 & 3.00 & 3.94 & 12.73 \\
2 & 98.0 & 94.0 & 107.0 & 113.6 & 2.98 & 4.02 & 16.27 \\
4 & 75.5 & 86.7 & 145.1 & 158.0 & 2.77 & 4.42 & 21.86 \\
8 & 83.3 & 85.9 & 185.7 & 208.4 & 2.86 & 4.24 & 20.99 \\
12 & 92.3 & 83.5 & 215.6 & 243.0 & 2.98 & 4.04 & 17.40 \\
16 & 91.5 & 82.9 & 244.2 & 280.6 & 2.99 & 4.01 & 16.54 \\
20 & 94.0 & 84.7 & 291.0 & 329.8 & 2.98 & 4.01 & 16.47 \\
30 & 93.0 & 84.2 & 460.6 & 524.6 & 2.99 & 4.04 & 16.61 \\
\addlinespace[1pt]
\multicolumn{8}{l}{\textbf{NADMM}} \\
1 & 98.0 & 99.7 & 63.9 & 58.2 & 2.98 & 3.99 & 12.19 \\
2 & 49.0 & 94.1 & 200.7 & 187.3 & 2.43 & 4.93 & 25.97 \\
4 & 13.0 & 93.5 & 263.6 & 342.9 & 1.66 & 6.02 & 42.61 \\
8 & 0.8 & 89.3 & 390.6 & 589.1 & 1.24 & 6.40 & 53.22 \\
12 & 1.3 & 92.5 & 583.5 & 823.7 & 1.14 & 7.87 & 53.93 \\
16 & 0.6 & 95.3 & 581.4 & 1063.7 & 0.75 & 7.07 & 55.19 \\
20 & 0.1 & 94.7 & 771.7 & 1321.4 & 0.51 & 6.50 & 56.45 \\
30 & 0.1 & 95.4 & 1022.2 & 2278.8 & 0.27 & 6.27 & 58.02 \\
\addlinespace[1pt]
\multicolumn{8}{l}{\textbf{MR-STORM}} \\
1 & 94.0 & 38.3 & 22.7 & 21.0 & 2.94 & 5.20 & 10.33 \\
2 & 91.0 & 39.8 & 22.8 & 20.4 & 2.99 & 5.05 & 11.27 \\
4 & 82.5 & 54.1 & 20.8 & 18.2 & 3.00 & 5.18 & 13.56 \\
8 & 82.5 & 53.2 & 24.7 & 21.9 & 2.99 & 5.28 & 17.03 \\
12 & 82.5 & 55.9 & 40.8 & 36.8 & 2.98 & 5.38 & 18.26 \\
16 & 82.6 & 56.3 & 59.2 & 53.3 & 2.99 & 5.32 & 17.18 \\
20 & 84.1 & 54.6 & 80.0 & 71.5 & 3.00 & 5.28 & 17.03 \\
30 & 79.3 & 52.5 & 133.1 & 123.3 & 2.98 & 5.44 & 18.45 \\
\addlinespace[1pt]
\multicolumn{8}{l}{\textbf{Gafur}} \\
1 & 78.0 & 78.4 & 29.7 & 16.7 & 2.60 & 7.79 & 15.92 \\
2 & 67.0 & 85.0 & 50.8 & 33.0 & 2.40 & 7.19 & 22.81 \\
4 & 65.5 & 77.8 & 82.8 & 62.1 & 2.31 & 7.37 & 26.50 \\
8 & 77.0 & 78.2 & 137.2 & 114.4 & 2.55 & 7.84 & 26.51 \\
12 & 72.7 & 77.4 & 171.5 & 148.5 & 2.48 & 8.26 & 26.89 \\
16 & 70.5 & 73.0 & 200.0 & 173.0 & 2.45 & 8.56 & 27.11 \\
20 & 65.2 & 72.9 & 223.5 & 195.4 & 2.24 & 7.61 & 26.04 \\
30 & 5.3 & 66.1 & 233.1 & 207.5 & 0.20 & 8.83 & 27.53 \\
\bottomrule
\end{tabular*}
\par\vspace{3pt}
\noindent{\fontsize{7.5}{8.6}\selectfont\CompactScalingNote\par}
\endgroup

%% file: tables/supp_scenario_results.tex
\providecommand{\CompactSettingHeader}{Setting}
\providecommand{\CompactSpacingLabel}{(i) Base spacing (m)}
\providecommand{\CompactSpeedLabel}{(ii) Obstacle speed (m/s)}
\providecommand{\CompactDropLabel}{(iii) Packet-loss rate}
\providecommand{\CompactDelayLabel}{(iv) Belief-update delay (cycles)}
\providecommand{\CompactNoiseLabel}{(v) Obstacle-noise s.d. (cm)}
\providecommand{\CompactTargetsLabel}{(vi) Targets per arm}
\providecommand{\CompactScenarioCaption}{Detailed scenario results for Fig.~\ref*{fig:scenario_ablation}.}
\providecommand{\CompactScenarioNote}{Each setting has 50 scenes and 200 arms. $\star$ denotes the shared DD-GBP reference. Units and metric definitions follow manuscript Tab.~\ref*{tab:batch_results}.}
\begingroup
\fontsize{8}{8.6}\selectfont
\setlength{\tabcolsep}{1.1pt}
\renewcommand{\arraystretch}{0.97}

\captionof{table}{\CompactScenarioCaption}\label{tab:supp_scenarios_compact}
\noindent\begin{tabular*}{\linewidth}{@{\extracolsep{\fill}}lrrrrrrr@{}}
\toprule
\CompactSettingHeader & Succ.$\uparrow$ & $\bar d_s\uparrow$ & $T_c\downarrow$ & $T_w\downarrow$ & $C_g\uparrow$ & $L_g\downarrow$ & $T_g\downarrow$ \\
 & (\%) & (mm) & (ms) & (ms) &  & (rad) & (s) \\
\midrule
\multicolumn{8}{l}{\textit{\CompactSpacingLabel}} \\
0.65 & 65.0 & 64.7 & 46.2 & 24.3 & 2.94 & 5.27 & 8.16 \\
0.7 & 82.0 & 71.6 & 35.8 & 19.7 & 2.99 & 4.55 & 6.61 \\
0.75 & 86.5 & 71.3 & 31.5 & 18.7 & 3.00 & 4.35 & 6.12 \\
0.8$^{\star}$ & 92.5 & 73.5 & 25.8 & 16.3 & 3.00 & 4.14 & 5.88 \\
\addlinespace[1pt]
\multicolumn{8}{l}{\textit{\CompactSpeedLabel}} \\
0.1 & 96.0 & 84.5 & 29.8 & 19.7 & 3.00 & 4.13 & 6.01 \\
0.15 & 93.5 & 79.2 & 40.0 & 26.3 & 3.00 & 4.13 & 5.91 \\
0.2$^{\star}$ & 92.5 & 73.5 & 25.8 & 16.3 & 3.00 & 4.14 & 5.88 \\
0.25 & 81.0 & 64.6 & 24.8 & 16.7 & 3.00 & 4.23 & 5.81 \\
0.3 & 55.0 & 57.4 & 29.8 & 20.4 & 3.00 & 4.32 & 5.97 \\
0.4 & 26.5 & 46.5 & 29.8 & 20.5 & 2.99 & 4.54 & 6.26 \\
\addlinespace[1pt]
\multicolumn{8}{l}{\textit{\CompactDropLabel}} \\
0$^{\star}$ & 92.5 & 73.5 & 25.8 & 16.3 & 3.00 & 4.14 & 5.88 \\
0.1 & 90.5 & 73.7 & 23.5 & 15.6 & 3.00 & 4.15 & 5.86 \\
0.3 & 89.0 & 73.2 & 23.2 & 15.7 & 3.00 & 4.18 & 5.94 \\
0.5 & 84.5 & 71.9 & 21.8 & 15.2 & 3.00 & 4.18 & 5.91 \\
0.7 & 87.0 & 72.8 & 21.6 & 14.9 & 3.00 & 4.14 & 5.72 \\
0.9 & 86.0 & 73.1 & 19.3 & 14.3 & 3.00 & 4.18 & 5.79 \\
\addlinespace[1pt]
\multicolumn{8}{l}{\textit{\CompactDelayLabel}} \\
0$^{\star}$ & 92.5 & 73.5 & 25.8 & 16.3 & 3.00 & 4.14 & 5.88 \\
1 & 85.0 & 72.8 & 24.9 & 16.6 & 3.00 & 4.19 & 5.83 \\
2 & 86.5 & 73.3 & 24.8 & 16.0 & 3.00 & 4.17 & 5.91 \\
3 & 88.5 & 72.9 & 24.1 & 16.0 & 3.00 & 4.14 & 5.77 \\
4 & 87.0 & 73.4 & 24.1 & 16.3 & 3.00 & 4.20 & 5.95 \\
5 & 89.0 & 73.5 & 25.3 & 16.4 & 3.00 & 4.17 & 5.81 \\
\addlinespace[1pt]
\multicolumn{8}{l}{\textit{\CompactNoiseLabel}} \\
0$^{\star}$ & 92.5 & 73.5 & 25.8 & 16.3 & 3.00 & 4.14 & 5.88 \\
1 & 86.5 & 72.2 & 23.3 & 16.0 & 3.00 & 4.15 & 5.86 \\
3 & 82.5 & 69.3 & 23.5 & 16.1 & 3.00 & 4.15 & 5.80 \\
5 & 77.5 & 68.1 & 24.1 & 16.1 & 2.99 & 4.19 & 6.05 \\
7 & 70.5 & 66.7 & 23.9 & 16.4 & 3.00 & 4.20 & 5.99 \\
9 & 59.0 & 60.3 & 23.5 & 16.5 & 3.00 & 4.23 & 6.00 \\
\addlinespace[1pt]
\multicolumn{8}{l}{\textit{\CompactTargetsLabel}} \\
1 & 93.5 & 74.1 & 27.0 & 14.8 & 0.96 & 3.30 & 4.55 \\
2 & 88.0 & 75.1 & 23.6 & 15.3 & 1.95 & 4.43 & 8.33 \\
3$^{\star}$ & 92.5 & 73.5 & 25.8 & 16.3 & 3.00 & 4.14 & 5.88 \\
4 & 86.0 & 72.4 & 22.9 & 15.6 & 4.00 & 4.24 & 5.87 \\
5 & 80.0 & 71.2 & 21.8 & 15.0 & 5.00 & 4.28 & 5.91 \\
\bottomrule
\end{tabular*}
\par\vspace{3pt}
\noindent{\fontsize{7.5}{8.6}\selectfont\CompactScenarioNote\par}
\endgroup

%% file: tables/supp_execution_diagnostics.tex
\begingroup
\centering
\caption{Recorded execution diagnostics: 50 scenes and 200 arms per configuration.}
\label{tab:supp_execution_diagnostics}\label{tab:supp_execution_tracking}
\label{tab:supp_execution_velocity}\label{tab:supp_execution_acceleration}
\begingroup\fontsize{9}{11}\selectfont
\setlength{\tabcolsep}{4pt}\renewcommand{\arraystretch}{1.10}
\begin{tabular*}{0.99\textwidth}{@{\extracolsep{\fill}}lrrrrrr@{}}
\toprule

Configuration & $e_q$ (aligned) & $e_q^{\rm raw}$ & $v_{\rm RMS}$ & $\max|\dot q|$ & $a_{\rm RMS}$ & $\max|\ddot q|$\\
 & (mrad) & (mrad) & (rad/s) & (rad/s) & (rad/s$^2$) & (rad/s$^2$)\\
\midrule
DD-GBP & $2.41\pm1.14$ & $29.04\pm2.77$ & $0.286\pm0.0264$ & 0.900 & $1.82\pm0.440$ & 53.2\\
Syn-GBP & $5.26\pm13.50$ & $31.72\pm8.04$ & $0.296\pm0.0320$ & 7.40 & $2.33\pm1.94$ & 760.7\\
GPMP2 & $15.81\pm23.00$ & $36.13\pm15.43$ & $0.280\pm0.0422$ & 6.10 & $4.14\pm3.29$ & 659.9\\
Gafur (NMPC) & $7.59\pm6.78$ & $7.54\pm6.75$ & $0.303\pm0.150$ & 4.99 & $14.5\pm16.6$ & 880.9\\
MR-STORM & $6.86\pm1.08$ & $9.65\pm1.32$ & $0.185\pm0.0236$ & 4.06 & $2.34\pm1.08$ & 527.9\\
DMPC & $4.53\pm2.53$ & $10.85\pm2.16$ & $0.102\pm0.0104$ & 3.98 & $2.23\pm0.529$ & 388.8\\
NADMM & $4.41\pm0.87$ & $12.17\pm3.04$ & $0.115\pm0.0269$ & 2.96 & $2.08\pm0.426$ & 380.4\\
\midrule
DD-GBP w/ prediction & $2.43\pm2.09$ & $29.20\pm2.99$ & $0.286\pm0.0261$ & 5.06 & $1.80\pm0.602$ & 607.5\\
w/o tree messages & $2.53\pm0.94$ & $29.09\pm2.53$ & $0.288\pm0.0246$ & 3.03 & $1.87\pm0.567$ & 511.3\\
w/o triggering & $1.98\pm0.83$ & $27.05\pm2.32$ & $0.268\pm0.0215$ & 1.90 & $1.78\pm0.427$ & 165.9\\
w/o adaptive $N$ & $4.60\pm2.17$ & $25.10\pm2.86$ & $0.246\pm0.0275$ & 5.62 & $2.34\pm0.773$ & 308.4\\
w/o subgraph & $2.76\pm4.96$ & $27.95\pm3.77$ & $0.275\pm0.0280$ & 16.0 & $1.84\pm0.953$ & 442.5\\
w/o GP scaling & $2.11\pm1.84$ & $27.55\pm2.71$ & $0.272\pm0.0252$ & 3.37 & $1.84\pm0.609$ & 358.5\\
w/o adaptive $\Delta t$ & $2.30\pm1.51$ & $30.72\pm2.43$ & $0.303\pm0.0231$ & 5.85 & $1.87\pm0.817$ & 666.5\\
w/o interpolation & $4.35\pm8.78$ & $29.81\pm5.01$ & $0.285\pm0.0267$ & 6.31 & $2.33\pm2.01$ & 888.1\\

\bottomrule
\end{tabular*}
\par\vspace{3pt}

{\footnotesize RMS: arm mean $\pm$ standard deviation; peaks: full-window maxima across 200 arms. Acceleration uses timestamp-aware velocity differences.}

\endgroup
\endgroup

%% file: collision_math.tex
\begingroup
\setlength{\parskip}{3.5pt}
\setlength{\abovedisplayskip}{6pt}\setlength{\belowdisplayskip}{6pt}
\setlength{\abovedisplayshortskip}{4pt}\setlength{\belowdisplayshortskip}{4pt}
\begin{minipage}[t]{0.485\textwidth}
\section*{S3. \HL{Collision-Factor Jacobians and Linearization Example}}
\renewcommand{\thesection}{S\arabic{section}}
\setcounter{section}{2}\refstepcounter{section}
\label{sec:supp_collision_jacobians}
\HL{For either residual in Eq.~}\HL{\eqref{eq:collision_residual}}\HL{, write} \HL{$d(\boldsymbol q)=d_c(\boldsymbol q)$}\HL{,} \HL{$\epsilon=\epsilon_c$}\HL{, and} \HL{$\boldsymbol{x}=[\boldsymbol q;\dot{\boldsymbol q}]$}\HL{. Let} \HL{\(n_c=\dim\boldsymbol q\)}\HL{, with} \HL{\(n_c=N_q\)} for \HL{one arm and} \HL{\(n_c=2N_q\)} \HL{for an inter-arm factor. Away from the activation boundary} \HL{$d(\boldsymbol q)=\epsilon$}\HL{, differentiation gives the full-state Jacobian}
\HL{\begin{equation}
\boldsymbol{J}_{\mathrm c}(\boldsymbol{x})=
\begin{cases}
\left[-\epsilon^{-1}\nabla_{\boldsymbol q}d(\boldsymbol q)^\top,\ \boldsymbol{0}_{1\times n_c}\right],
&d(\boldsymbol q)<\epsilon,\\
\boldsymbol{0}_{1\times 2n_c},&d(\boldsymbol q)>\epsilon.
\end{cases}
\label{eq:supp_collision_jacobian}
\end{equation}}
\HL{At the boundary, the inactive branch supplies the zero-Jacobian convention. The smoothed distance is $d(\boldsymbol q)=-\alpha^{-1}\log\sum_\ell\exp[-\alpha d_\ell(\boldsymbol q)]$, $\alpha>0$. Differentiating this smoothed distance yields}
\HL{\begin{equation}
\nabla_{\boldsymbol q}d(\boldsymbol q)=
\frac{\sum_\ell e^{-\alpha d_\ell(\boldsymbol q)}\boldsymbol J_{\mathrm p,\ell}(\boldsymbol q)^\top\boldsymbol n_\ell(\boldsymbol q)}{\sum_\ell e^{-\alpha d_\ell(\boldsymbol q)}}.
\label{eq:supp_softmin_gradient}
\end{equation}}
\HL{Here} \HL{$\nabla_{\boldsymbol q}d_\ell=\boldsymbol J_{\mathrm p,\ell}^{\top}\boldsymbol n_\ell$}\HL{, where} \HL{$\boldsymbol J_{\mathrm p,\ell}$} \HL{is the closest-point geometric Jacobian and} \HL{$\boldsymbol n_\ell$} \HL{the corresponding unit witness direction. For inter-arm distances,} \HL{$\boldsymbol q$} \HL{stacks both arms' joints and the relative-point Jacobian is} \HL{$[\boldsymbol J_{\mathrm p,k,\ell},-\boldsymbol J_{\mathrm p,j,\ell}]$}\HL{, with the witness direction oriented from arm} \HL{$j$} \HL{to arm} \HL{$k$}.

\HL{For the GP interpolation $\boldsymbol x(\tau)=\mathbf L_i(\tau)\boldsymbol x^i+\mathbf L_{i+1}(\tau)\boldsymbol x^{i+1}$, $t_i<\tau<t_{i+1}$, the obstacle residual has endpoint Jacobians $\boldsymbol J_{\mathrm c}(\boldsymbol x(\tau))\mathbf L_i(\tau)$ and $\boldsymbol J_{\mathrm c}(\boldsymbol x(\tau))\mathbf L_{i+1}(\tau)$. Although Eq.~\eqref{eq:supp_collision_jacobian} has zero columns for the instantaneous velocity, the endpoint velocity columns are generally nonzero because the GP interpolation depends on both endpoint velocities.}

\end{minipage}\hfill
\begin{minipage}[t]{0.485\textwidth}
\begingroup\color{black}
\subsection*{A numerical collision-factor linearization}\label{sec:supp_linearization_example}
Consider a planar two-link arm, used here only to make the calculation reproducible, with link lengths $l_1=0.4$~m and $l_2=0.3$~m. A collision sphere centered at the endpoint has combined radius $r_{\mathrm a}+r_{\mathrm o}=0.05$~m with a spherical obstacle at $\boldsymbol c=(0.5,0.3)$~m. Define
\[
\boldsymbol p(\boldsymbol q)=
\begin{bmatrix}l_1\cos q_1+l_2\cos(q_1+q_2)\\l_1\sin q_1+l_2\sin(q_1+q_2)\end{bmatrix},
\]\[
d(\boldsymbol q)=\|\boldsymbol p(\boldsymbol q)-\boldsymbol c\|-0.05 .
\]
At $\boldsymbol q^0=(0,\pi/2)$, $\boldsymbol p^0=(0.4,0.3)$, $d^0=0.05$~m, and $\boldsymbol n=(-1,0)^\top$. For $\epsilon=0.10$~m the active residual is $h=(\epsilon-d)/\epsilon=0.5$. The point and residual Jacobians are
\[
J_{\mathrm p}=
\begin{bmatrix}-0.3&-0.3\\0.4&0\end{bmatrix},
\]\[
\nabla_{\boldsymbol q}d=J_{\mathrm p}^{\top}\boldsymbol n=
\begin{bmatrix}0.3\\0.3\end{bmatrix},
\]\[
J_h=\begin{bmatrix}-3&-3&0&0\end{bmatrix}
\]
for $\boldsymbol x=[q_1,q_2,\dot q_1,\dot q_2]^\top$. Hence
$h(\boldsymbol x^0+\delta\boldsymbol x)\simeq0.5-3\delta q_1-3\delta q_2$.
For $\delta\boldsymbol q=(0.01,-0.005)$~rad, the linear value is $0.485$ and direct nonlinear evaluation gives $0.484014$.

\endgroup

\end{minipage}
\endgroup

%% file: main.bbl
\begin{thebibliography}{10}
\providecommand{\url}[1]{#1}
\csname url@samestyle\endcsname
\providecommand{\newblock}{\relax}
\providecommand{\bibinfo}[2]{#2}
\providecommand{\BIBentrySTDinterwordspacing}{\spaceskip=0pt\relax}
\providecommand{\BIBentryALTinterwordstretchfactor}{4}
\providecommand{\BIBentryALTinterwordspacing}{\spaceskip=\fontdimen2\font plus
\BIBentryALTinterwordstretchfactor\fontdimen3\font minus \fontdimen4\font\relax}
\providecommand{\BIBforeignlanguage}[2]{{%
\expandafter\ifx\csname l@#1\endcsname\relax
\typeout{** WARNING: IEEEtran.bst: No hyphenation pattern has been}%
\typeout{** loaded for the language `#1'. Using the pattern for}%
\typeout{** the default language instead.}%
\else
\language=\csname l@#1\endcsname
\fi
#2}}
\providecommand{\BIBdecl}{\relax}
\BIBdecl

\bibitem{park2024constrained}
S.~Park, S.~Jeon, and J.~Park, ``A constrained motion planning method exploiting learned latent space for high-dimensional state and constraint spaces*,'' \emph{IEEE/ASME Trans. Mechatronics}, vol.~29, no.~4, pp. 3001--3009, 2024.

\bibitem{WANG2025102849}
I.-L. Wang and T.-H. Wang, ``Efficient routing in robotic movable fulfillment systems with integer programming: A rolling horizon and heuristic approach,'' \emph{Robot. Comput.-Integr. Manuf.}, vol.~91, p. 102849, 2025.

\bibitem{zhou2019real}
Y.~Zhou, H.~Hu, Y.~Liu, S.-W. Lin, and Z.~Ding, ``A real-time and fully distributed approach to motion planning for multirobot systems,'' \emph{IEEE Trans. Syst., Man, Cybern., Syst.}, vol.~49, no.~12, pp. 2636--2650, 2019.

\bibitem{deWitte2022learning}
S.~De~Witte, T.~Van~Hauwermeiren, T.~Lefebvre, and G.~Crevecoeur, ``Learning to cooperate: A hierarchical cooperative dual robot arm approach for underactuated pick-and-placing,'' \emph{IEEE/ASME Trans. Mechatronics}, vol.~27, no.~4, pp. 1964--1972, 2022.

\bibitem{dong2024bimanual}
Y.~Dong, Z.~Yu, X.~Chen, X.~Zhu, C.~Wang, P.~Gergondet, and Q.~Huang, ``Bimanual continuous steering wheel turning by a dual-arm robot,'' \emph{IEEE/ASME Trans. Mechatronics}, vol.~29, no.~3, pp. 1773--1784, 2024.

\bibitem{zhao2025adaptive}
X.~Zhao, Y.~Song, and H.~Ye, ``Adaptive distributed event-triggered cooperative manipulation of multiple manipulators under partial time-interval error constraints,'' \emph{IEEE Trans. Syst., Man, Cybern., Syst.}, vol.~55, no.~12, pp. 9309--9323, 2025.

\bibitem{wagner2011m}
G.~Wagner and H.~Choset, ``M*: A complete multirobot path planning algorithm with performance bounds,'' in \emph{Proc. IEEE/RSJ IROS}, 2011, pp. 3260--3267.

\bibitem{chen2025cimap}
B.~Chen, H.~Zhang, F.~Zhang, Y.~Jiang, W.~He, C.~Yang, and Y.~Wang, ``Cimap: A high-performance motion planning algorithm for robotic manipulators in complex environments using clearance inference network,'' \emph{IEEE Trans. Syst., Man, Cybern., Syst.}, vol.~55, no.~9, pp. 6383--6396, 2025.

\bibitem{solovey2016finding}
K.~Solovey, O.~Salzman, and D.~Halperin, ``Finding a needle in an exponential haystack: Discrete rrt for exploration of implicit roadmaps in multi-robot motion planning,'' \emph{Int. J. Robot. Res.}, vol.~35, no.~5, pp. 501--513, 2016.

\bibitem{JIN2026103923}
Z.~Jin, J.~Yu, Y.~Liang, Y.~Wang, Z.~Wang, and C.~Hu, ``Co-dosp: A hierarchical optimization-based motion planner for multi-robot manipulation in confined and task-constrained workspace,'' \emph{Adv. Eng. Inform.}, vol.~69, p. 103923, 2026.

\bibitem{zhang2023online}
S.~Zhang and F.~Pecora, ``Online and scalable motion coordination for multiple robot manipulators in shared workspaces,'' \emph{IEEE Trans. Autom. Sci. Eng.}, 2023.

\bibitem{WEI2025102850}
L.~Wei, Y.~Wang, Y.~Hu, T.~L. Lam, and Y.~Wei, ``Online dual robot–human collaboration trajectory generation by convex optimization,'' \emph{Robot. Comput.-Integr. Manuf.}, vol.~91, p. 102850, 2025.

\bibitem{boyd2011distributed}
S.~Boyd, N.~Parikh, E.~Chu, B.~Peleato, and J.~Eckstein, ``Distributed optimization and statistical learning via the alternating direction method of multipliers,'' \emph{Found. Trends Mach. Learn.}, vol.~3, no.~1, pp. 1--122, 2011.

\bibitem{ferrantiDistributedNonlinearTrajectory2023}
L.~Ferranti, L.~Lyons, R.~R. Negenborn, T.~Keviczky, and J.~{Alonso-Mora}, ``Distributed nonlinear trajectory optimization for multi-robot motion planning,'' \emph{IEEE Trans. Control Syst. Technol.}, vol.~31, no.~2, pp. 809--824, Mar. 2023.

\bibitem{schwartingParallelAutonomyAutomated2017}
W.~Schwarting, J.~Alonso-Mora, L.~Pauli, S.~Karaman, and D.~Rus, ``Parallel autonomy in automated vehicles: Safe motion generation with minimal intervention,'' in \emph{Proc. IEEE ICRA}, 2017, pp. 1928--1935.

\bibitem{gafur2022dynamic}
N.~Gafur, G.~Kanagalingam, A.~Wagner, and M.~Ruskowski, ``Dynamic collision and deadlock avoidance for multiple robotic manipulators,'' \emph{IEEE Access}, vol.~10, pp. 55\,766--55\,781, 2022.

\bibitem{jiangDeployingTenThousand}
H.~Jiang, Y.~Wang, R.~Veerapaneni, T.~H. Duhan, G.~A. Sartoretti, and J.~Li, ``Deploying ten thousand robots: Scalable imitation learning for lifelong multi-agent path finding,'' in \emph{Proc. IEEE ICRA}, 2025.

\bibitem{aljalbout2022clascoordinatingmultirobot}
\BIBentryALTinterwordspacing
E.~Aljalbout, M.~Karl, and P.~van~der Smagt, ``Clas: Coordinating multi-robot manipulation with central latent action spaces,'' 2022. [Online]. Available: \url{https://arxiv.org/abs/2211.15824}
\BIBentrySTDinterwordspacing

\bibitem{lee2025multiagent}
H.~Lee, C.~Luo, and H.~Jung, ``Multi-agent deep reinforcement learning for collision-free posture control of multi-manipulators in shared workspaces,'' \emph{Sensors}, vol.~25, no.~22, 2025.

\bibitem{laiRoboBalletPlanningMultirobot2025}
M.~Lai, K.~Go, Z.~Li, T.~Kr{\"o}ger, S.~Schaal, K.~Allen, and J.~Scholz, ``Roboballet: Planning for multirobot reaching with graph neural networks and reinforcement learning,'' \emph{Sci. Robot.}, 2025.

\bibitem{evron2026mrstorm}
D.~Evron, E.~Goldsztejn, and R.~I. Brafman, ``Decoupled {MPPI}-based multi-arm motion planning,'' arXiv:2602.10114, 2026.

\bibitem{mukadamContinuoustimeGaussianProcess2018}
M.~Mukadam, J.~Dong, X.~Yan, F.~Dellaert, and B.~Boots, ``Continuous-time gaussian process motion planning via probabilistic inference,'' \emph{Int. J. Robot. Res.}, vol.~37, no.~11, pp. 1319--1340, 2018.

\bibitem{dellaertFactorGraphsExploiting2021}
F.~Dellaert, ``Factor graphs: Exploiting structure in robotics,'' \emph{Annu. Rev. Control Robot. Auton. Syst.}, vol.~4, no.~1, pp. 141--166, 2021.

\bibitem{guo2021formation}
S.~Guo, B.~Liu, S.~Zhang, J.~Guo, and C.~Wang, ``Continuous-time {Gaussian} process trajectory generation for multi-robot formation via probabilistic inference,'' in \emph{Proc. IEEE/RSJ Int. Conf. Intell. Robots Syst. (IROS)}, 2021, pp. 9247--9253.

\bibitem{wang2022multimanipulator}
W.~Wang and B.~Hu, ``Multi-manipulator motion planning based on {Gaussian} process with probabilistic inference,'' in \emph{Proc. WRC Symp. Adv. Robot. Autom. (WRC SARA)}, 2022, pp. 221--226.

\bibitem{bishop2006pattern}
C.~M. Bishop, \emph{Pattern Recognition and Machine Learning}.\hskip 1em plus 0.5em minus 0.4em\relax Springer, 2006.

\bibitem{suConvergenceConditionsGaussian2015}
Q.~Su and Y.-C. Wu, ``On convergence conditions of gaussian belief propagation,'' \emph{IEEE Trans. Signal Process.}, vol.~63, no.~5, pp. 1144--1155, 2015.

\bibitem{duConvergenceAnalysisDistributed2018}
\BIBentryALTinterwordspacing
J.~Du, S.~Ma, Y.-C. Wu, S.~Kar, and J.~M.~F. Moura, ``Convergence analysis of distributed inference with vector-valued gaussian belief propagation,'' \emph{J. Mach. Learn. Res.}, vol.~18, no. 172, pp. 1--38, 2018. [Online]. Available: \url{https://jmlr.org/papers/v18/16-556.html}
\BIBentrySTDinterwordspacing

\bibitem{ortizVisualIntroductionGaussian2021}
J.~Ortiz, T.~Evans, and A.~J. Davison, ``A visual introduction to gaussian belief propagation,'' 2021.

\bibitem{davisonFutureMapping2Gaussian2022}
A.~J. Davison and J.~Ortiz, ``Futuremapping 2: Gaussian belief propagation for spatial ai,'' 2022.

\bibitem{muraiRobotWebDistributed2024}
R.~Murai, J.~Ortiz, S.~Saeedi, P.~H.~J. Kelly, and A.~J. Davison, ``A robot web for distributed many-device localisation,'' \emph{IEEE Trans. Robot.}, vol.~40, pp. 121--138, 2024.

\bibitem{patwardhanDistributingCollaborativeMultiRobot2023}
A.~Patwardhan, R.~Murai, and A.~J. Davison, ``Distributing collaborative multi-robot planning with gaussian belief propagation,'' \emph{IEEE Robot. Autom. Lett.}, vol.~8, no.~2, pp. 552--559, 2023.

\bibitem{bari2023fastrelinearization}
S.~Bari, V.~Gabler, and D.~Wollherr, ``Probabilistic inference-based robot motion planning via {Gaussian} belief propagation,'' \emph{IEEE Robot. Autom. Lett.}, vol.~8, no.~8, pp. 5156--5163, 2023.

\bibitem{prados2025distributed}
A.~Prados, G.~Espinoza, L.~Moreno, and R.~Barber, ``Coordination of learned decoupled dual-arm tasks through {Gaussian} belief propagation,'' in \emph{Proc. IEEE/RSJ IROS}, 2025, pp. 15\,917--15\,924.

\bibitem{jin2026distributedSafetyAware}
Z.~Jin, F.~Dai, Y.~Liang, Z.~Sun, Z.~Wang, and C.~Hu, ``Distributed safety-aware motion planning for dual-arm systems via {Gaussian} belief propagation with application to catenary dropper assembly,'' \emph{IEEE Robot. Autom. Lett.}, vol.~11, no.~11, pp. 12\,791--12\,798, 2026.

\bibitem{zhangSchurComplement2005}
F.~Zhang, Ed., \emph{The Schur Complement and Its Applications}.\hskip 1em plus 0.5em minus 0.4em\relax Springer, 2005.

\bibitem{weissCorrectnessBelief2001}
Y.~Weiss and W.~T. Freeman, ``Correctness of belief propagation in gaussian graphical models of arbitrary topology,'' \emph{Neural Comput.}, vol.~13, no.~10, pp. 2173--2200, 2001.

\bibitem{malioutovWalkSumsBelief2006}
D.~M. Malioutov, J.~K. Johnson, and A.~S. Willsky, ``Walk-sums and belief propagation in {G}aussian graphical models,'' \emph{J. Mach. Learn. Res.}, vol.~7, pp. 2031--2064, 2006.

\bibitem{moallemiConvergenceMinSumConvex2010}
C.~C. Moallemi and B.~Van~Roy, ``Convergence of the min-sum algorithm for convex optimization,'' \emph{IEEE Trans. Inf. Theory}, vol.~56, no.~4, pp. 2041--2050, 2010.

\bibitem{kamperConvergenceGaBPNodes2019}
F.~Kamper, S.~J. Steel, and J.~A. du~Preez, ``On the convergence of gaussian belief propagation with nodes of arbitrary size,'' \emph{J. Mach. Learn. Res.}, vol.~20, no. 165, pp. 1--37, 2019.

\bibitem{dellaert2012gtsam}
\BIBentryALTinterwordspacing
F.~Dellaert, ``Factor graphs and {GTSAM}: A hands-on introduction,'' Georgia Institute of Technology, Tech. Rep. GT-RIM-CP\&R-2012-002, Sep. 2012. [Online]. Available: \url{https://hdl.handle.net/1853/45226}
\BIBentrySTDinterwordspacing

\end{thebibliography}
